\documentclass{article}

\usepackage{arxiv}

\usepackage[utf8]{inputenc} 
\usepackage[T1]{fontenc}    
\usepackage{hyperref}       
\usepackage{url}            
\usepackage{booktabs}       
\usepackage{amsfonts}       
\usepackage{nicefrac}       
\usepackage{microtype}      
\usepackage{lipsum}
\usepackage{graphicx}
\graphicspath{ {./images/} }
\usepackage{tabularx}
\usepackage{siunitx}        
\usepackage{xcolor}

\newcolumntype{Y}{>{\centering\arraybackslash}X}
\usepackage{wrapfig}
\usepackage{booktabs}
\usepackage{tabularx}
\usepackage{siunitx}
\usepackage{xcolor}
\usepackage{colortbl}
\usepackage{adjustbox}
\usepackage{graphicx}
\usepackage{amssymb}
\usepackage{amsmath}
\usepackage{url}
\usepackage{algorithm}
\usepackage{algorithmic}
\usepackage{subcaption}
\usepackage{xcolor}
\usepackage{amssymb}
\usepackage{breqn}
\usepackage{ragged2e}
\usepackage{wrapfig}
\usepackage{graphicx}
\usepackage{placeins}
\usepackage{makecell}
\usepackage{cite}

\title{A Self supervised learning framework for imbalanced medical imaging datasets}

\author{
 Yash Kumar Sharma \\
  Artificial Intelligence lab \\ 
  School of Computer \& Information Sciences\\ 
  University of Hyderabad \\ 
  Hyderabad 500046, India \\
  \texttt{21mcpc11@uohyd.ac.in} \\
   \And
 Charan Ramtej Kodi \\
  Artificial Intelligence lab \\ 
  School of Computer \& Information Sciences\\ 
  University of Hyderabad \\ 
  Hyderabad 500046, India \\
  \texttt{21mcpc10@uohyd.ac.in} \\
  \And
 Vineet Padmanabhan \\
  Artificial Intelligence lab \\ 
  School of Computer \& Information Sciences\\ 
  University of Hyderabad \\ 
  Hyderabad 500046, India \\
  \texttt{vineetnair@uohyd.ac.in} \\
}

\begin{document}
\maketitle
\begin{abstract}
Two problems often plague medical imaging analysis: 1) Non-availability of large quantities of labeled training data, and 2) Dealing with imbalanced data, i.e., abundant data are available for frequent classes, whereas data are highly limited for the rare class. Self supervised learning (SSL) methods have been proposed to deal with the first problem to a certain extent, but the issue of investigating the robustness of SSL to imbalanced data has rarely been addressed in the domain of medical image classification. In this work, we make the following contributions: 1) The MIMV method proposed by us in an earlier work is extended with a new \emph{augmentation} strategy to construct asymmetric multi-image, multi-view (AMIMV) pairs to address both data scarcity and dataset imbalance in medical image classification. 2) We carry out a data analysis to evaluate the robustness of AMIMV under varying degrees of class imbalance in medical imaging .
%
%
3) We evaluate eight representative SSL methods in 11 medical imaging datasets (MedMNIST) under long-tailed distributions and limited supervision. Our experimental results on the MedMNIST dataset show an improvement of 4.25\% on retinaMNIST, 1.88\% on tissueMNIST, and 3.1\% on DermaMNIST.

\end{abstract}
\section{Introduction}

Deep learning techniques have demonstrated significant achievements in the analysis of medical images for tasks such as disease classification, screening, and diagnosis in a supervised setting~\cite{AljuaidA22survey}. Supervised learning methods are constrained by the scarcity of annotated data and the naturally imbalanced characteristics of medical datasets~\cite{azizi23nature}. Acquiring expert annotations is both expensive and labor-intensive, and numerous rare conditions are poorly represented, leading to long-tailed class distributions that impair model performance in the medical context. Long-tailed learning occurs when abundant data are available for frequent classes, whereas data are highly limited for the rare class. 
Self-supervised learning (SSL), on the other hand, does not rely on labeled datasets for supervisory signals and often generates implicit labels by leveraging the underlying structure in the data. SSL has achieved impressive results with regard to image data representation without labels in benchmark datasets such as Imagenet \cite{russakovsky2015imagenet} and Cifar \cite{krizhevsky2009learning}.  
%
Recent works on SSL has also demonstrated that compared to fully supervised models, architectures that take advantage of self-supervised pretraining are more robust to class imbalance and long-tailed learning\cite{yang2020rethinking} \cite{liu2021self} \cite{lin2023frequency} \cite{bai2023effectiveness}\\ \cite{jiang2021self}\cite{kukleva2023temperature}. 
These methods makes use of in-domain (ID) data samples to balance the minority class to boost the long-tailed learning performance of SSL. 
Other methods address the long-tailed problem by strengthening minority features through sampling or reweighting techniques \cite{liu2021self}. There is also a work that addresses the problem of data set imbalance in SSL using a prototypical rebalancing strategy \cite{lin2023frequency}.  
Of these, except for~\cite{bai2023effectiveness}, all other works address the data imbalance problem by sampling with extra in-domain data that can rebalance the minority class. Moreover,  all the above mentioned works deal with synthetically imbalanced data (using exponential/pareto distribution). 

Despite these observations, the robustness of SSL methods for medical image analysis in long-tailed settings has not been systematically evaluated, 
Existing approaches frequently rely on ImageNet-pretrained encoders or auxiliary out-of-distribution data, limiting their applicability to real-world medical scenarios. The only recent work we are aware of that applies self supervised learning on imbalanced medical imaging datasets is that of [Andreas Espis)]. In this work, the main idea is to contrast supervised learning(SL) with that of the traditional multi-view SSL models (VicReg and MAE)  on medical image datasets. We differ from this work in the following ways:

\begin{itemize}
    \item We make use of the  multi-image multi-view framework as outlined in our earlier work \cite{sharma2025maximally} where the traditional single image multi-view assumption is replaced with a novel combination of invertible (normalized) and non-invertible (augmented) transformations to build training views designed for medical imaging classification.
 \item We perform an imbalance analysis of all MedMNIST 2D datasets using both relative (Imbalance Ratio, Coefficient of Variation, Normalized Entropy and Gini Index) and absolute (Rare Class Ratio) measures, and systematically categorize the datasets into \texttt{fairly balanced}, \texttt{partially imbalanced}, and \texttt{severely imbalanced}. This analysis provides a basis for evaluating the robustness of self-supervised learning under varying degrees of class imbalance in medical imaging, which is quite novel.
 \item Since medical image datasets are quite sensitive to perturbations, we introduce a novel augmentation strategy that is suitable for MedMNIST dataset.
 
 \item Our experimental results on the MedMNIST dataset show an improvement of 4.25\% on retinaMNIST, 1.88\% on tissueMNIST and 3.1\% on DermaMNIST. These three datasets are specifically chosen due to their higher degree of imbalance. Compared to \cite{espis2025comparative}, the improvement over VicReg in these three datasets is 7.6\% in retinaMNIST, 5.36\% in tissueMNIST, and 4.79\% in DermaMNIST.
    
\end{itemize}

\section{Related Work}

Recent advances in self-supervised learning have led to the development of various contrastive and non-contrastive frameworks for representation learning. Contrastive approaches such as SimCLR\cite{chen2020simple},MoCov3\cite{chen2021empirical} and ReSSL\cite{zheng2021ressl} rely on instance discrimination with large numbers of negative samples, while non-contrastive methods including BYOL~\cite{grill2020bootstrap},DINO~\cite{caron2021emerging},VICReg~\cite{bardes2021vicreg},NNCLR~\cite{dwibedi2021little} and Barlow Twins~\cite{zbontar2021barlow} avoid explicit negatives through architectural or regularization-based constraints. These methods show strong performance on natural-image benchmarks, but are primarily evaluated under balanced data assumptions. 
In this work, we adopt these representative SSL approaches as baselines to systematically evaluate robustness under class imbalance in medical imaging.

Previous self-supervised learning evaluations focus primarily on natural image datasets \cite{goyal2019scaling,marks2025closer}, resulting in a lack of systematic studies on medical imaging. Existing medical SSL analysis often relies on ImageNet-1K pretrained encoders without further medical-domain adaptation \cite{doerrich2025rethinking}, limiting the ability to capture clinically significant and important  features. Some works demonstrate that self-supervised pretraining can outperform supervised baselines, such as in pathology imaging \cite{kang2023benchmarking}, but their scope is restricted to a single modality. Consequently, the field lacks a unified evaluation in various medical tasks and imaging conditions.\cite{wolf2023self} compared contrastive methods such as MoCo, SwAV, BYOL and a masked autoencoder in large-scale unlabeled CT data, demonstrating consistent performance gains over random initialization when fine-tuned in downstream medical tasks, including OrganSMNIST.\\\cite{azizi2021big} further showed that self-supervised pretraining in unlabeled medical images combined with ImageNet initialization significantly outperforms supervised pretraining in dermatology and chest X-ray datasets, even under severe class imbalance.
These works highlight the benefits of SSL pretraining in medical imaging, but they primarily rely on different datasets for pretraining and fine-tuning, whereas we have a more constrained scenario wherein the same dataset is used for both pretraining and fine-tuning.

\cite{tian2021divide} analyzes SSL on uncurated data by implicitly separating dominant and rare visual patterns. \cite{assran2022hidden} shows that many SSL objectives impose a uniform clustering prior, which degrades performance when class distributions are long-tailed. In contrast, medical imaging datasets exhibit an inherent real-world class imbalance. We focus on representation learning directly on imbalanced medical datasets, without assuming balanced or curated data distributions. 

\begin{figure}[t]
    \centering
    \begin{minipage}[b]{0.45\linewidth}
        \centering
        \includegraphics[width=4cm, height=5cm]{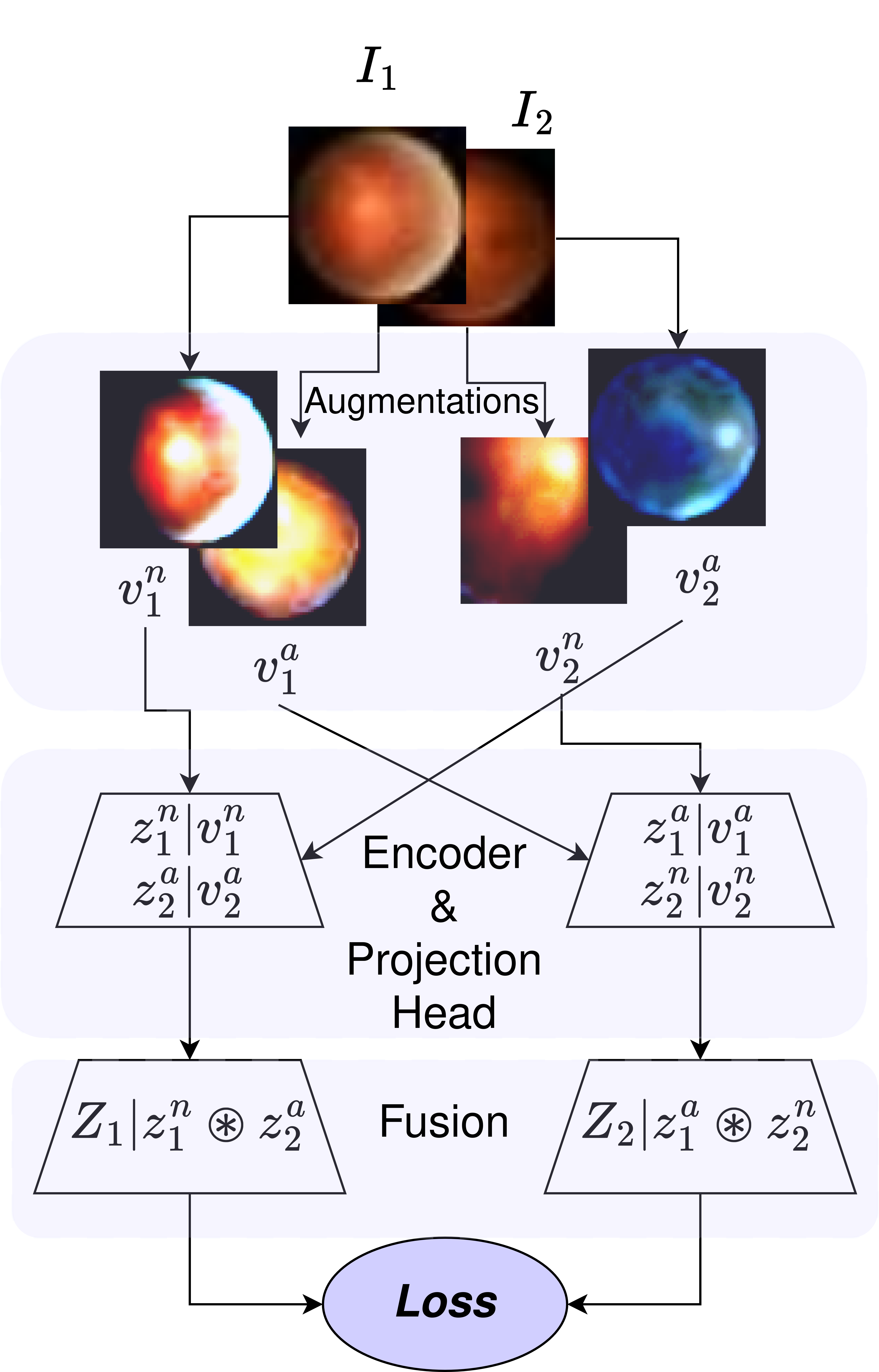}
        \caption*{(a) AMIMV-SSL illustration}
    \end{minipage}
    \hfill
    \begin{minipage}[b]{0.45\linewidth}
        \centering
        \includegraphics[width=3cm, height=4cm]{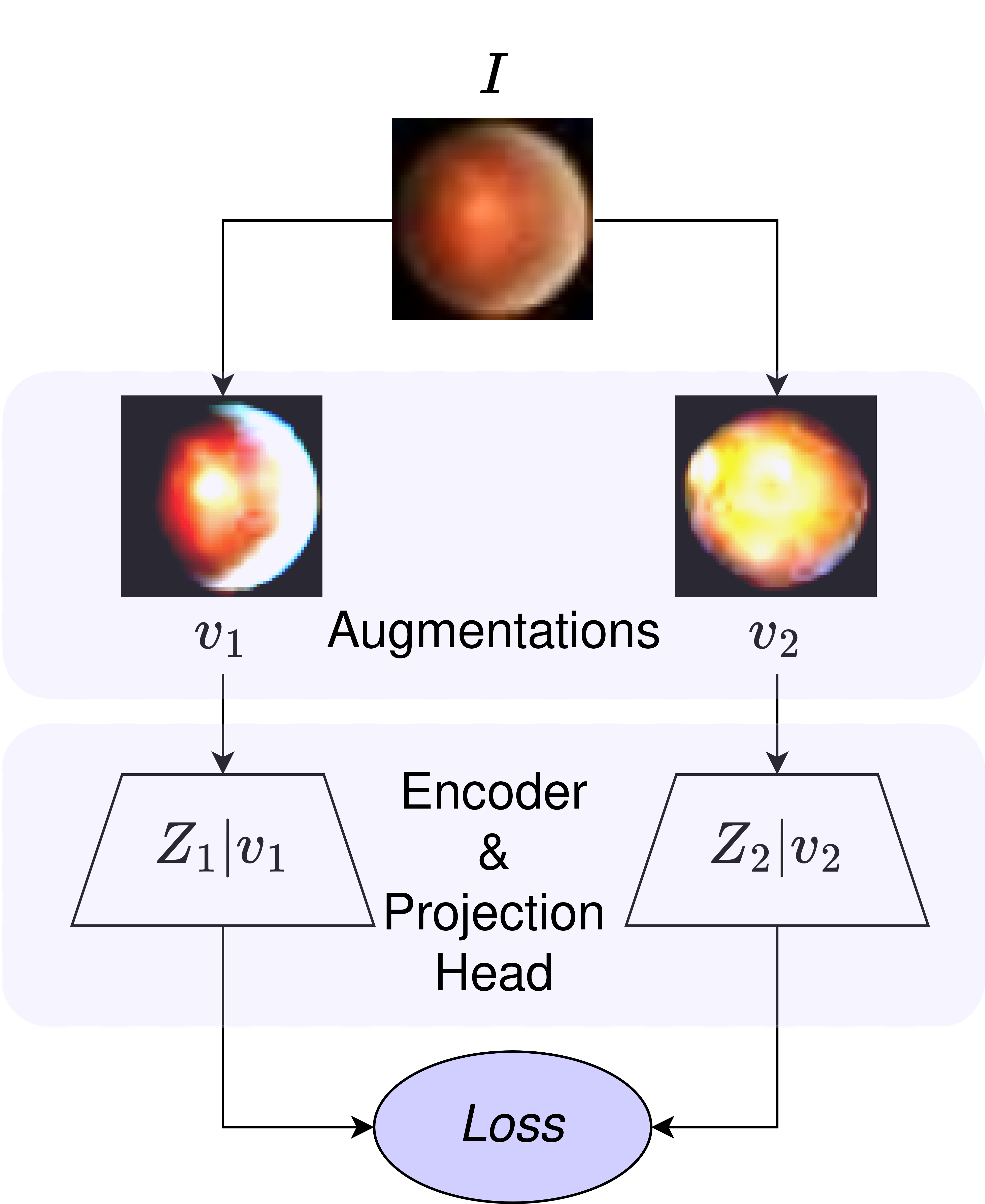}
        \caption*{(b) Single Image Multi-view illustration}
    \end{minipage}
    \caption{Comparison between AMIMV-SSL (asymmetric multi-image view self-supervised learning) and single multi-view learning.}
    \label{fig:mv_comparison}
\end{figure}

\section{Methodology}
Most self supervised learning frameworks~\cite{he2020momentum,khosla2020supervised} ~\cite{jiang2021self,ren2022simple} rely on NT-Xent loss or its variants which fall under the family of \emph{multi-view} self supervised learning. 
The basic assumption in multi-view SSL is to generate two or more views for each data sample by using augmentations so that the semantic information shared between the views remains as intact as that of the original image (as shown in Fig.\ref{fig:mv_comparison}(b)). 
In this work, we analyze the NT-Xent loss in the context of SimCLR framework~\cite{chen2020simple}, which belongs to the family of multi-view SSL and serves as a representative baseline for medical image pretraining on MedMNIST. NT-Xent computes pairwise similarity between two augmented views using cosine similarity, as defined in Equation~\ref{eq:nt-xent}.
\begin{align}
    \mathcal{L}_{i,j} &= -log\frac{exp(sim(z_i,z_j)/\tau)}{\sum_{k=1}^n \mathsf{1}_{\{i \neq k\}}exp(sim(z_i,z_k)/\tau)} \label{eq:nt-xent}
\end{align}
where $z_i$ and $z_j$ denote the latent representations of two augmented views from the same MedMNIST image or from different images within the batch, and $z_k$ represents the remaining latent representations excluding the positive pair $(z_i, z_j)$.

%
Figure~\ref{fig:mv_comparison} contrasts AMIMV-SSL with conventional single-image multi-view self-supervised learning\footnote{We coin the term multi-view as single image multi-view to distinguish with our terminology of multi-image multi-view}. Unlike standard SSL, which constructs positive pairs using multiple augmentations of the same image, AMIMV-SSL leverages asymmetric multi-image multiviews by combining normalized and augmented representations across distinct images. This design reduces redundancy and improves robustness in long-tailed medical data distributions. Given an anchor image and a corresponding counterpart image, we generate two views for each image: a normalized view ($v_1^n$,$v_2^n$) and an augmented view ($v_1^a$,$v_2^a$). For the anchor image, these are denoted as $v_1^n$ and $v_1^a$, while for the counterpart image they are denoted as $v_2^n$ and $v_2^a$, respectively. The normalized views preserve semantic and structural information, whereas the augmented views introduce controlled variability. The normalized view of the anchor image and the augmented view of the counterpart image are  passed through an encoder and a projection head and vise versa, which in turn leads to latent representations $z_1^n, z_1^a$, $z_2^n, z_2^a$. 

To jointly model inter-image and intra-image relationships, AMIMV-SSL forms fused representations of asymmetric cross-images. Specifically, the normalized representation of the anchor image is fused with the augmented representation of the counterpart image to form $Z_1 = z_1^n \circledast z_2^a,$ while the augmented representation of the anchor image is fused with the normalized representation of the counterpart image to form $Z_2 = z_1^a \circledast z_2^n,$ where $\circledast$ denotes the fusion operation. The fused representations $Z_1$ and $Z_2$ are then used to compute the contrastive loss. This asymmetric multi-image fusion enforces both \textbf{inter-image similarity} between the anchor and counterpart images and \textbf{intra-image consistency} across normalized and augmented views. By pairing one normalized view with one augmented view, the framework reduces the redundancy introduced by aggressive augmentations while retaining clinically meaningful information, which is crucial for medical image classification.

Consequently, the total loss over the pairs of representation is expressed as shown in \ref{eq:MIMV_loss}.This loss function serves as the NT-Xent loss function, utilizing a compact representation to distinguish between positive and negative samples.
\begin{align}
    \mathcal{L}_{i,j} = -log\frac{e^{\left(sim\left((z_A^n \circledast z_C^a)_i\;(z_A^a \circledast z_C^n)_j\right)/\tau\right)}}{\sum_{k=1}^{2N} \mathsf{1}_{\;\{i \neq k\}}e^{\left(sim\left((z_A^n \circledast z_C^a)_i\;(z_A^a \circledast z_C^n)_k\right)/\tau\right)} \label{eq:MIMV_loss}}
\end{align}

\begin{algorithm}
\caption{AMIMV-SSL: }
\begin{algorithmic}
\STATE \textbf{Input: } Batch size N, Normalized Images $v_1^n,v_2^n,$ Augmented Image $v_1^a, v_2^a,$ $encoder_q$, $encoder_k$,
\FOR{batch in train_loader}
    \STATE $v_1^n,v^2_n,v_1^a,v_2^a$ = batch
    \STATE $z_1^n,z_2^a$ = $encoder_q(v_1^n)$, $encoder_q(v_2^a)$
    \STATE \textbf{with} no_grad():
        \STATE \quad $momentum\_update\_encoder_k()$
        \STATE \quad $z^1_a,z_2^n$ = $encoder_k(v_1^a)$, $encoder_k(v_C^n)$
    \STATE $\mathcal{L}_{i,j} = -log\frac{exp(sim(z_i,z_j)/\tau)}{\sum_{k=1}^n \mathsf{1}_{\{i \neq k\}}exp(sim(z_i,z_k)/\tau)} $

\ENDFOR
\STATE return $\mathcal{L}$
\end{algorithmic}
\label{alg:MIMV_algorithm}
\end{algorithm}

Algorithm~\ref{alg:MIMV_algorithm} outlines the AMIMV-SSL pretraining procedure. We make use of the momentum learning~\cite{moco-improved} to update the weights of our encoders in which $encoder_{k}$ is the exponential moving average version of $encoder_{q}$'s parameters while $encoder_{q}$ is used to update the gradient by back propagation.


\section{Experiments}
\subsection{Dataset}
We assess our approach using the 2D subset of the MedMNIST benchmark \cite{yang2023medmnist}, which is a standardized set of lightweight biomedical image datasets covering various medical modalities, including pathology, microscopy, fundus imaging, ultrasound, and CT. MedMNIST provides uniformly preprocessed images with consistent spatial resolution and standardized train-validation-test splits, enabling fair and reproducible comparisons across various learning algorithms. In this study, we focus on 2D datasets, as our proposed AMIMV-SSL framework is tailored for contrastive representation learning specifically on image-level data. The chosen 2D datasets display different levels of class imbalance and varying dataset sizes, making them ideal for examining robustness against long-tailed distributions in the context of self-supervised pretraining.

\subsection{Dataset Analysis and Imbalance Characterization}

\begin{table*}[t]
\centering
\scriptsize
\setlength{\tabcolsep}{10pt}
\renewcommand{\arraystretch}{1.2} 
\caption{Class imbalance analysis of MedMNIST 2D datasets using relative (Imbalance Ratio, Coefficient of Variation, Normalized Entropy, Gini Index) and absolute (Rare Class Ratio) measures. Datasets are categorized as Fairly Balanced (FB), Partially Imbalanced (PI) and Imbalanced (I), considering both distribution skewness and minority class sufficiency.}
\label{tableimbalance}
\begin{tabular}{l|c|c|c|c|c|c}
\toprule
\textbf{Dataset}
& \textbf{IR} 
& \textbf{CV}
& \textbf{NE} 
& \textbf{GI}
& \textbf{RCR}
& \textbf{FB/PI/I} \\

\midrule
PathMNIST        &1.63  &0.16  &0.99  &0.08  &8.76  &FB \\
BloodMNIST       &2.74  &0.41  &0.96  &0.21  &7.09  &FB \\
OCTMNIST         &5.93  &0.75  &0.83  &0.35  &7.95  &FB \\
BreastMNIST      &2.71  &0.65  &0.84  &0.23  &26.92 &PI \\
PneumoniaMNIST   &2.87  &0.68  &0.82  &0.24  &25.78 &PI \\
OrganAMNIST      &4.54  &0.47  &0.95  &0.24  &3.92  &PI \\
OrganCMNIST      &5.01  &0.56  &0.94  &0.25  &4.58  &PI \\
OrganSMNIST      &5.64  &0.66  &0.93  &0.30  &4.40  &PI \\
RetinaMNIST      &7.36  &0.74  &0.87  &0.34  &6.11  &I \\
TissueMNIST      &9.04  &0.83  &0.86  &0.40  &3.54  &I \\
DermaMNIST       &58.66 &1.65  &0.58  &0.64  &1.14  &I \\
\bottomrule
\end{tabular}
\end{table*}

Table.\ref{tableimbalance} outlines the characteristics of class imbalance present in the MedMNIST 2D datasets using both relative metrics (Imbalance Ratio, Coefficient of Variation, Normalized Entropy, Gini Index) and absolute measurements (Rare Class Ratio). According to these metrics, PathMNIST, BloodMNIST, and OCTMNIST demonstrate low skewness in distribution, nearly uniform entropy, low Gini coefficients, and adequate representation of minority classes, thus classifying them as relatively balanced. Consequently, these datasets do not significantly challenge imbalance-aware learning strategies, and the variations in the performance on these datasets reflect overall representation quality 
In contrast, the other datasets fall into categories of partial or severe imbalance, displaying significant class skew and a limited number of minority samples. Since the AMIMV-SSL framework we propose is specifically designed to address representation bias and the dominance of head classes in imbalanced scenarios, we exclude fairly balanced datasets from our analysis of imbalance-specific results and focus our evaluation on partially and severely imbalanced datasets, where imbalance-aware self-supervised learning proves both essential and beneficial.

\subsection{Implementation Details}
AMIMV-SSL uses a unique augmentation strategy, distinct from the traditional approach of taking two augmented views of the same source image. AMIMV-SSL considers one normalized view and one augmented view of a source. It employs z-score normalization, using the mean and standard deviation of the dataset for one view. This strategy offers two advantages over traditional multi-view approaches. First, normalization ensures that all information remains intact in one view. Second, by using the augmented version of the image, our model learns invariant features.
Medical images are highly sensitive to augmentation and can lose information even with slight changes. Considering this, we adopt a novel augmented approach for second view, which comprises ColorJitter, RandomResizedCrop, RandomHorizontalFlip, and finally RandomGaussianBlur.For ColorJitter, we use brightness, contrast, and saturation value as 0.1, hue as 0.01 with p=0.8. RandomResizedCrop makes use of an image size of 64 with a scale of  (0.2, 1.0). RandomHorizontalFlip use  p=0.5 while RandomGaussianBlur makes use of kernel size (3,3) and sigma = (0.1,1.0). Throughout the experiments we resize the images to 64 and uses 3*3 kernel.

We pretrain a ResNet-50 encoder with a three-layer MLP projection head (hidden dimension 512, output dimension 128). We evaluate representation quality using alignment and uniformity metrics. We train the model with SGD using a cosine learning rate schedule and linear warm-up (10\% of epochs) starting from $1e-4$. The base learning rate is 0.75 and is scaled with batch size as 
\begin{equation}
    lr=0.75 * \frac{batch\_size}{256}    
\end{equation}
All models are trained from scratch for 400 epochs with a 128 batch size on each dataset.


\textbf{Linear Evaluation:} After Pretraining we consider trained backbone (Encoder\_q) only with out projection head.Over the Frozen RESNET-50, we trained one linear layer for the classification task. First, we extract the features of each MEDMNIST dataset and train . To optimize this task we use the learning rate of 0.005 with the AdamW optimizer. In this training process,we also make use of a cosine learning rate without warmup. We trained for 100 epochs and a batch size of 128.

\subsection{Representation Analysis}
\begin{figure*}[h!]
    \centering

    \subfloat[PathMNIST]{%
        \includegraphics[width=0.3\linewidth]{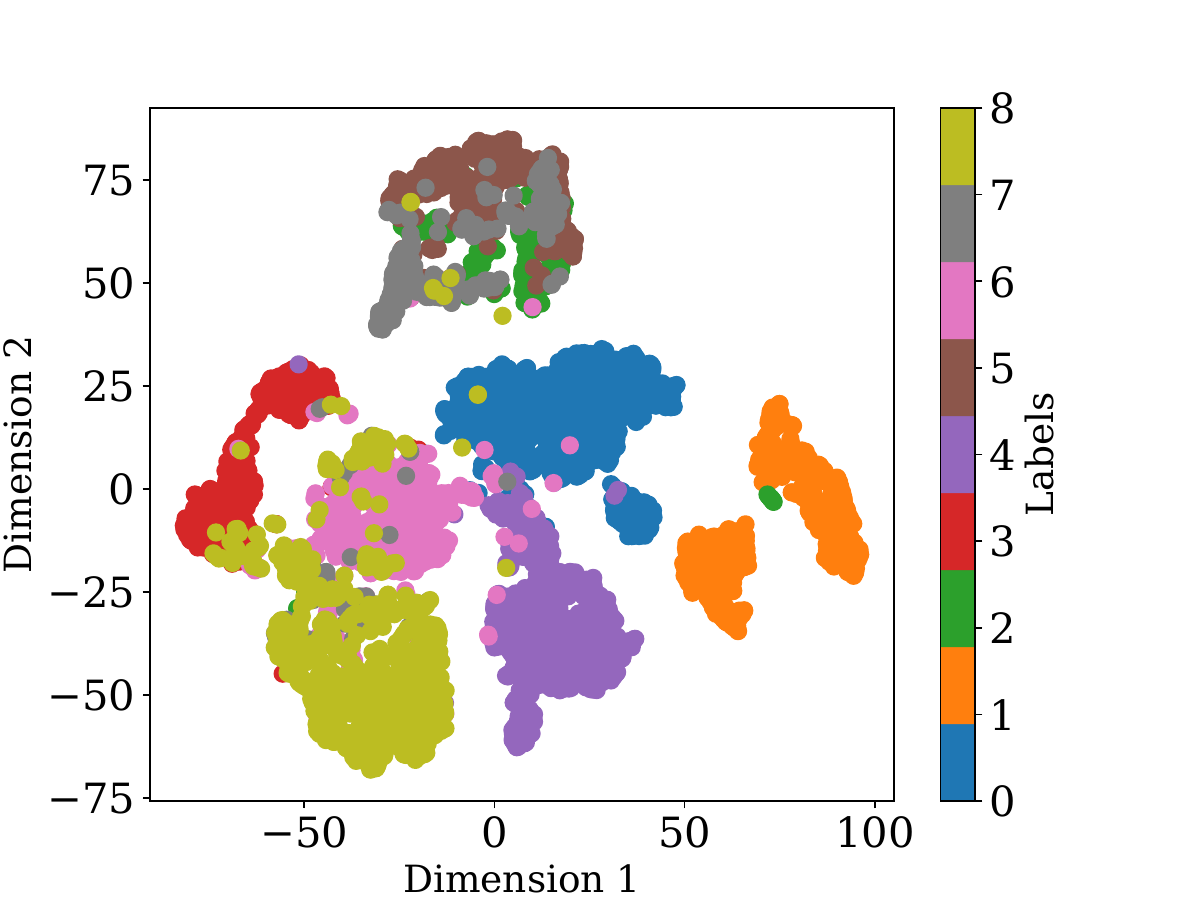}
    }\hfill
    \subfloat[BloodMNIST]{%
        \includegraphics[width=0.3\linewidth]{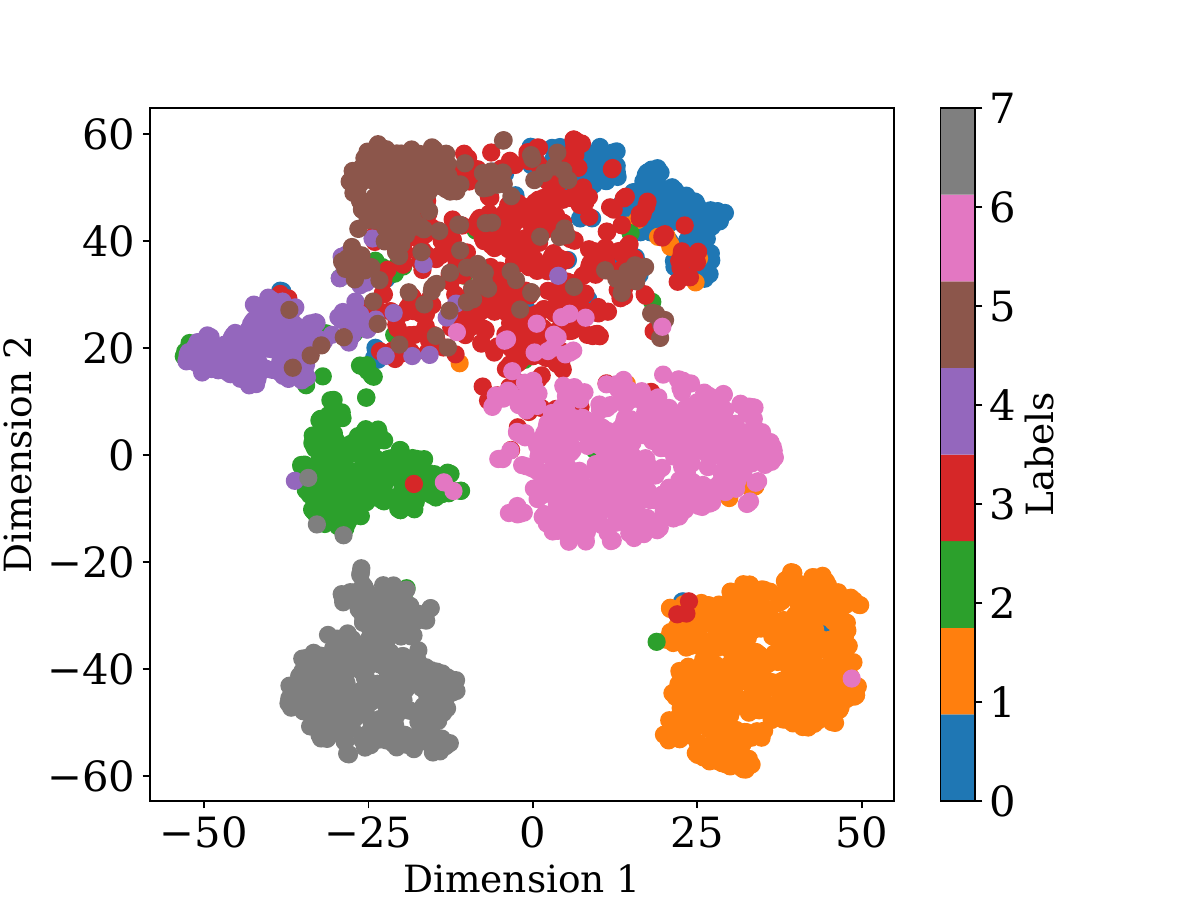}
    }\hfill
    \subfloat[OCTMNIST]{%
        \includegraphics[width=0.3\linewidth]{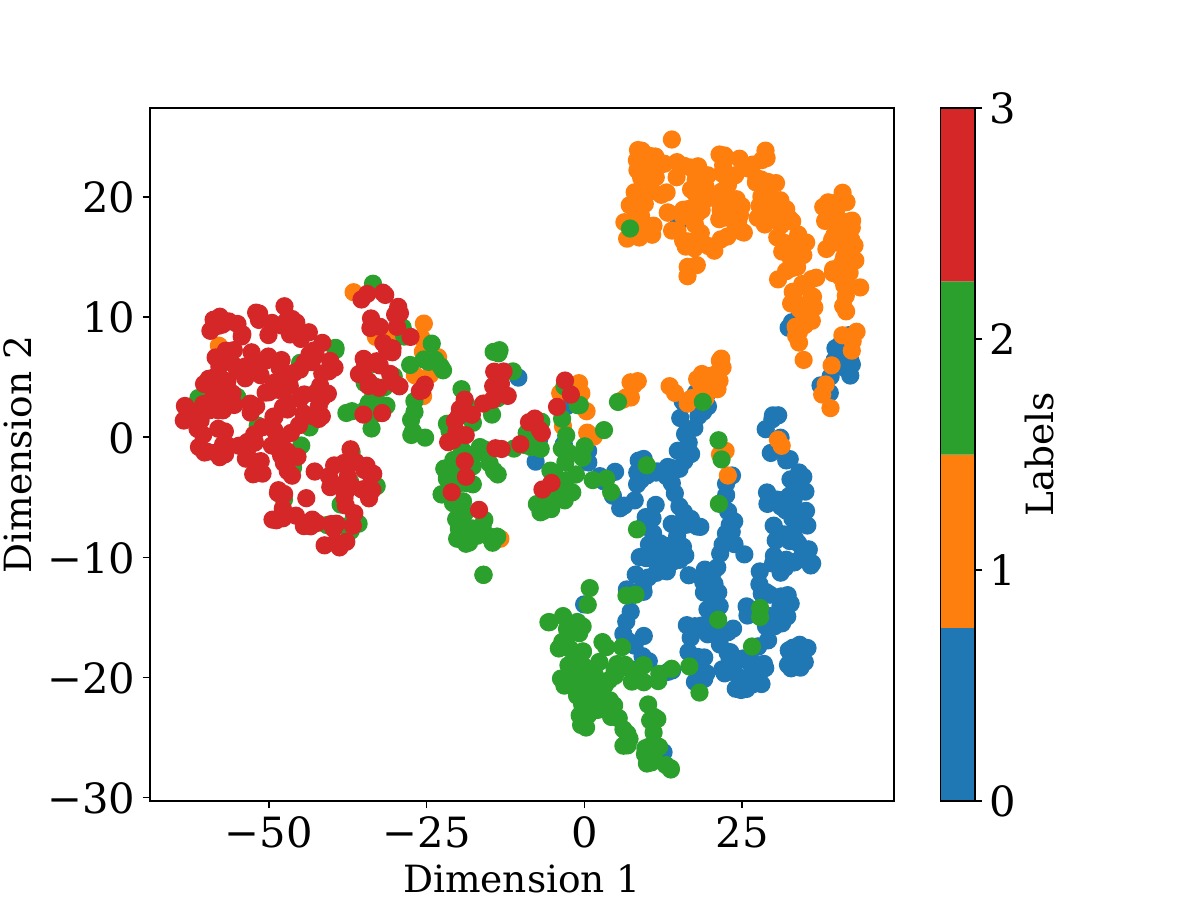}
    }

    \vspace{0.25cm}

    \subfloat[BreastMNIST]{%
        \includegraphics[width=0.3\linewidth]{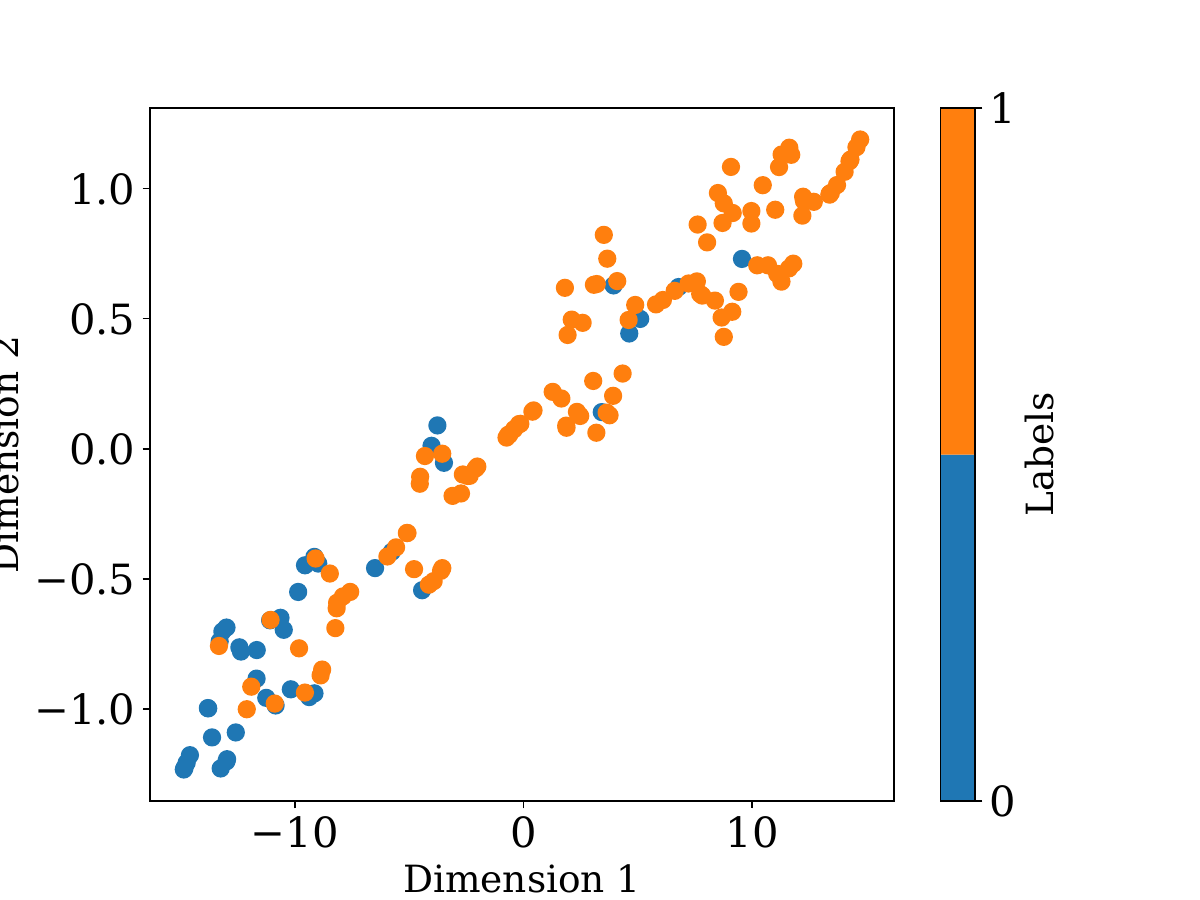}
    }\hfill
    \subfloat[PneumoniaMNIST]{%
        \includegraphics[width=0.3\linewidth]{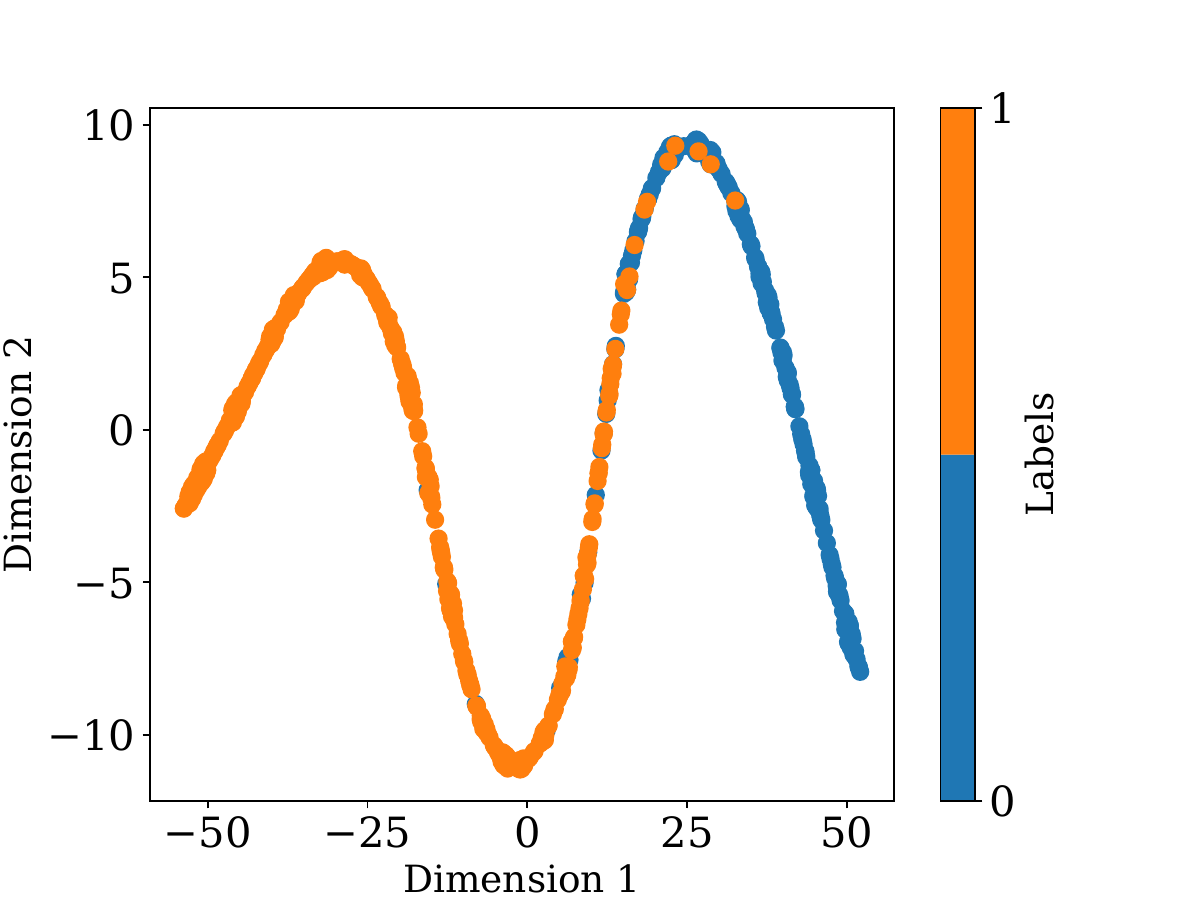}
    }\hfill
    \subfloat[OrganAMNIST]{%
        \includegraphics[width=0.3\linewidth]{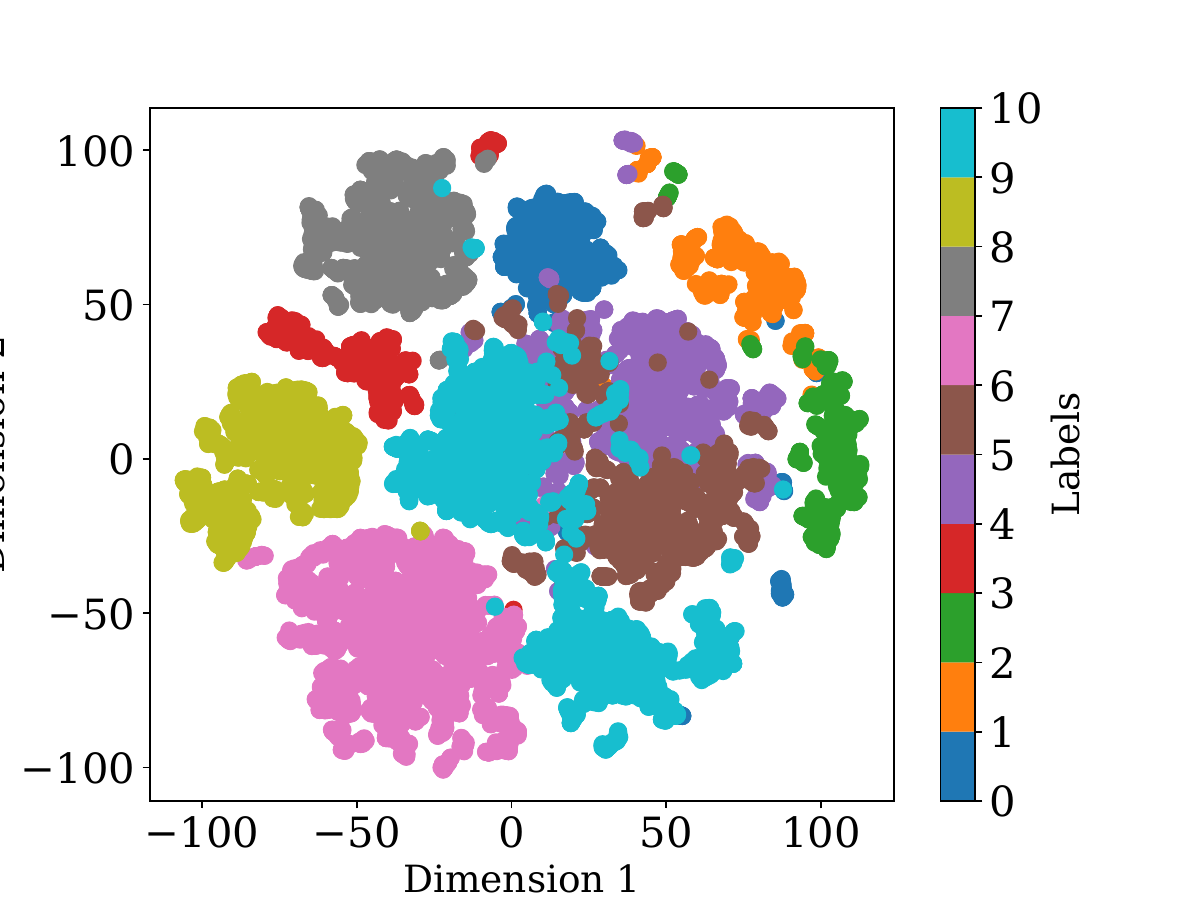}
    }

    \vspace{0.25cm}

    \subfloat[OrganCMNIST]{%
        \includegraphics[width=0.3\linewidth]{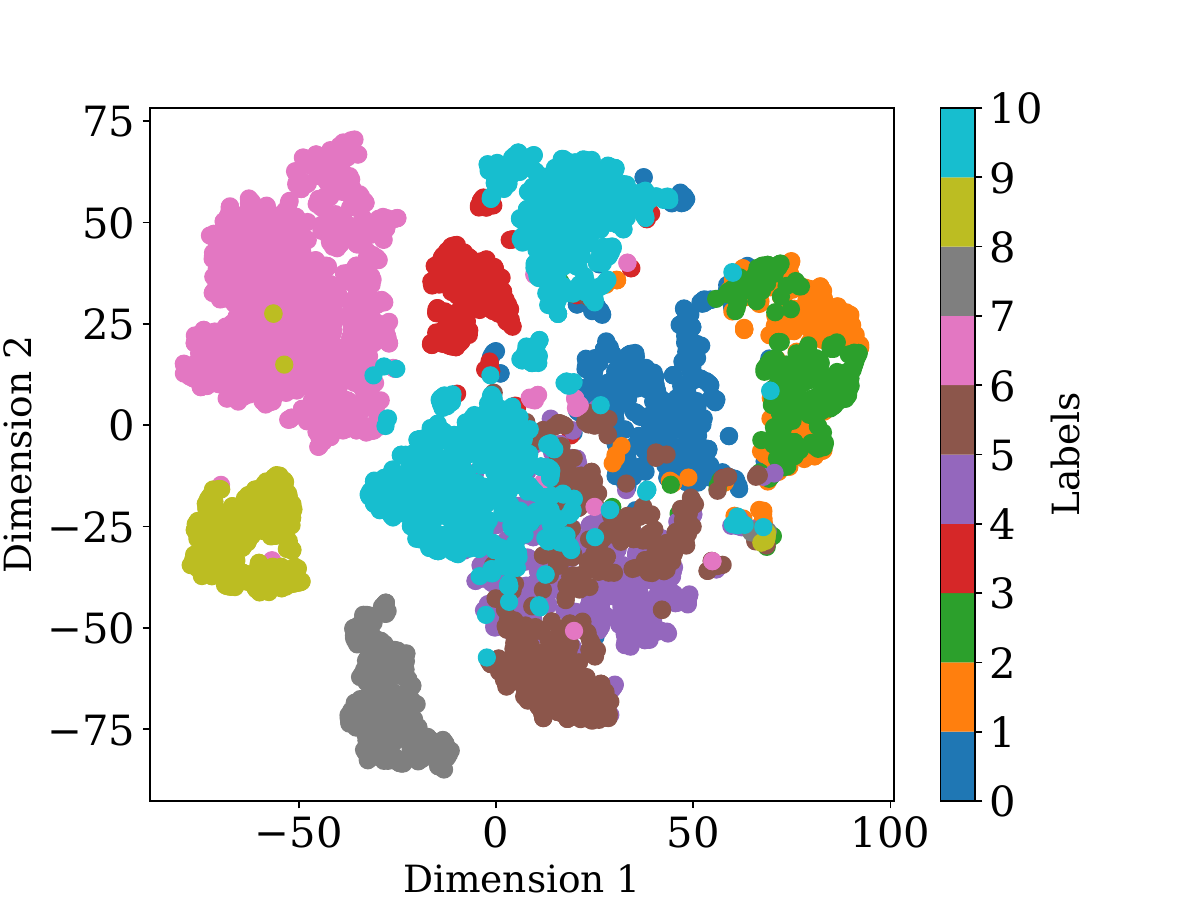}
    }\hfill
    \subfloat[OrganSMNIST]{%
        \includegraphics[width=0.3\linewidth]{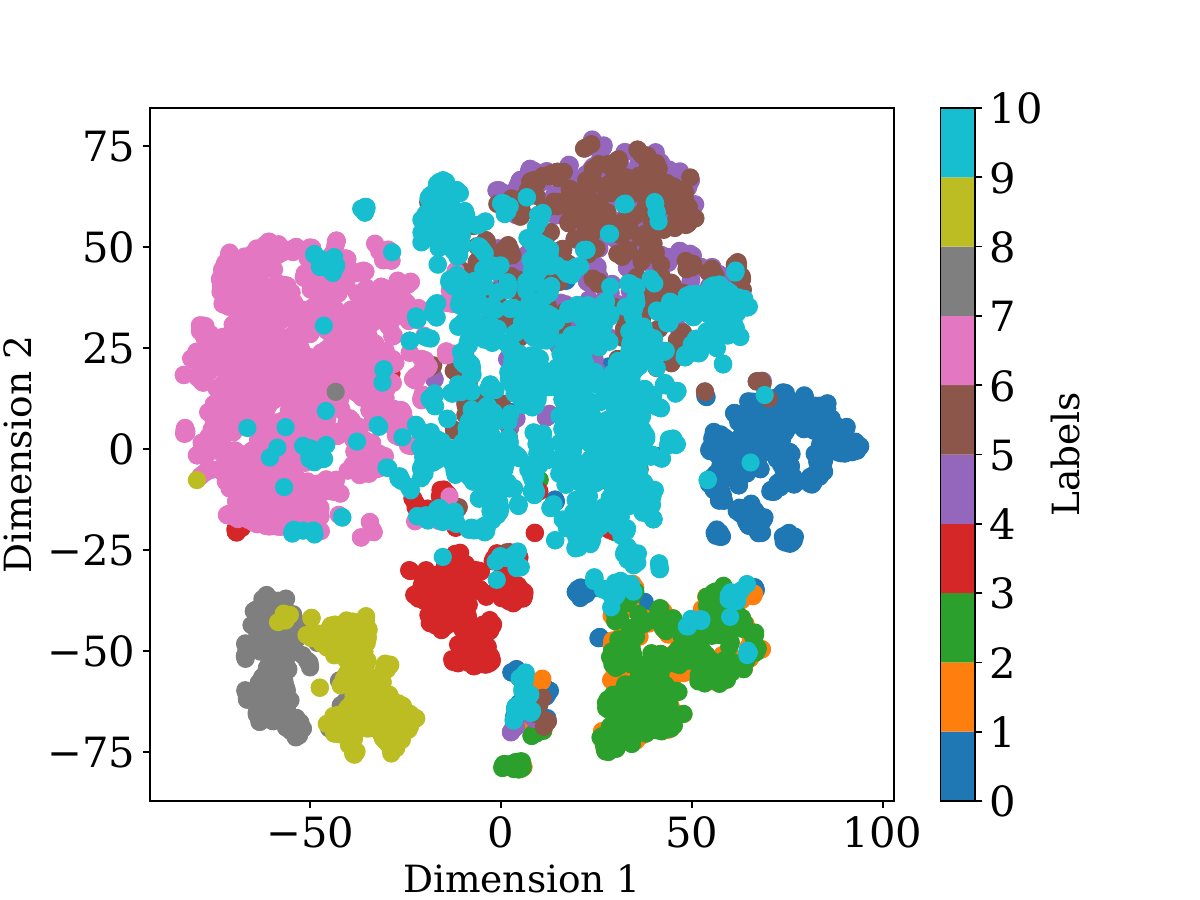}
    }\hfill
    \subfloat[RetinaMNIST]{%
        \includegraphics[width=0.3\linewidth]{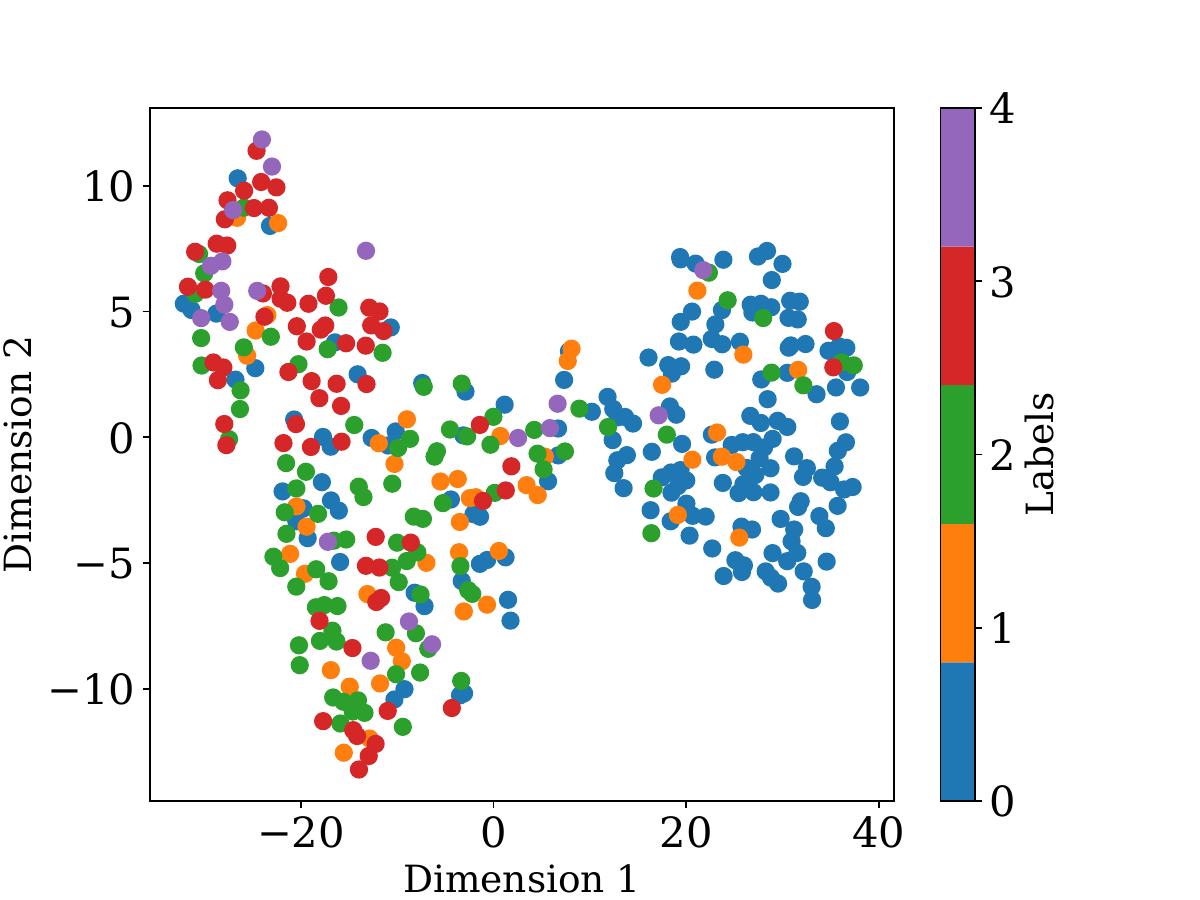}
    }

    \vspace{0.25cm}

    \subfloat[TissueMNIST]{%
        \includegraphics[width=0.3\linewidth]{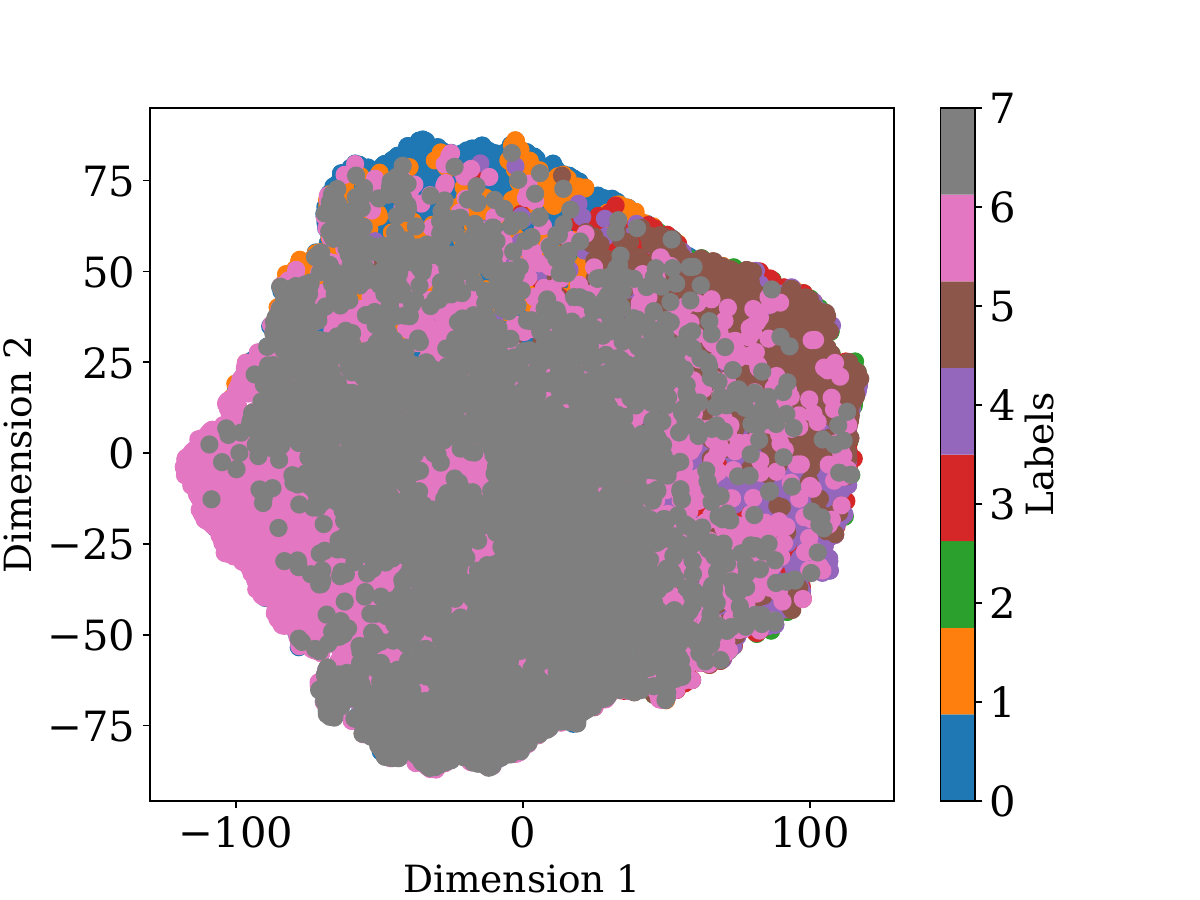}
    }\hfill
    \subfloat[DermaMNIST]{%
        \includegraphics[width=0.3\linewidth]{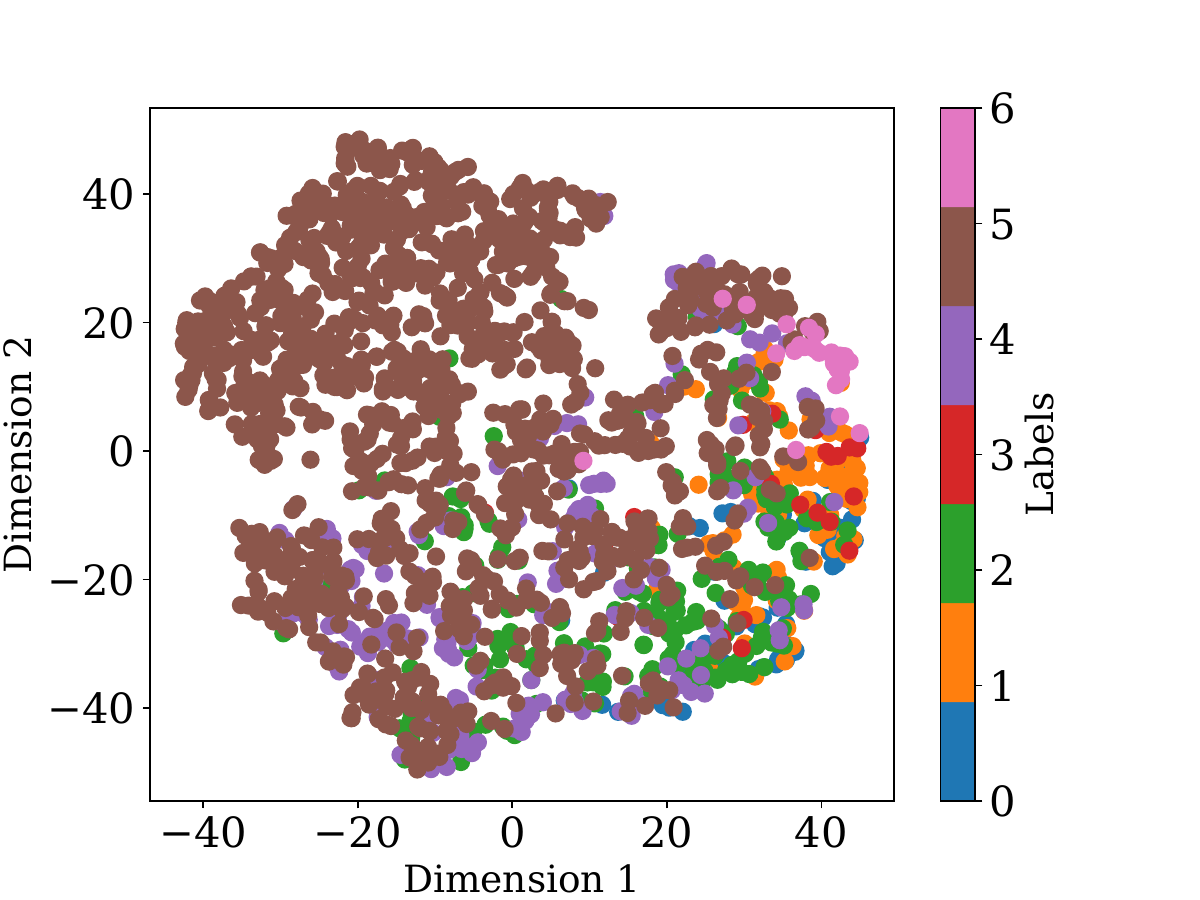}
    }

    \caption{t-SNE visualizations showing encoder representations that are learned by AMIMV-SSL across eleven MedMNIST datasets. Each subplot represents a distinct dataset, with colors representing ground truth class labels. The tight and distinctly separated clusters indicate strong and discriminative representation learning across various medical imaging modalities and imbalanced class distributions.}
    \label{fig:tsne_all}
\end{figure*}

Figure.\ref{fig:tsne_all} demonstrates the t-SNE visualizations of feature embeddings learned by AMIMV-SSL across eleven datasets from the MEDMNIST, including PathMNIST, BloodMNIST, OCTMNIST, BreastMNIST, PneumoniaMNIST, OrganAMNIST, OrganCMNIST, OrganSMNIST, RetinaMNIST, TissueMNIST, and DermaMNIST.Across these different modalities, the learned embeddings show tightly packed and distinctly separate clusters for each class. Significantly, classes with a limited number of samples are still clearly identifiable rather than merging into larger class regions, demonstrating robustness to long-tailed data distributions. This pattern is consistently seen in both binary and multi-class datasets, including PneumoniaMNIST, BreastMNIST, and organ-based datasets. Moreover, the reduced overlap between clusters suggests that our method AMIMV-SSL suppresses redundant and majority-biased features while preserving semantically meaningful information. These visualizations qualitatively demonstrate that AMIMV-SSL learns structured and discriminative representations that generalize across heterogeneous medical image modalities without relying on labeled supervision during pretraining.

\section{Results}
We compare our method with representative state-of-the-art self-supervised learning approaches, including MoCo v3 \cite{chen2021empirical}, SimCLR \cite{chen2020simple}, DINO \cite{caron2021emerging}, ReSSL \cite{zheng2021ressl}, BYOL \cite{grill2020bootstrap}, VICReg \cite{bardes2021vicreg}, NNCLR \cite{dwibedi2021little}, and Barlow Twins \cite{zbontar2021barlow}. All methods are evaluated using a ResNet-50 backbone to ensure a fair comparison.

\begin{table*}[t]
\centering
\scriptsize
\setlength{\tabcolsep}{1.3pt}
\renewcommand{\arraystretch}{1.15} 
\caption{Mean accuracy scores of random initialization for self-supervised training using ResNet-50 across different methods and datasets.}
\label{MIMVcomparision1}
\begin{tabular}{l|c|c|c|c|c|c|c|c|c}
\toprule
\textbf{Dataset}
& \textbf{MoCoV3} 
& \textbf{SimCLR}
& \textbf{DINO} 
& \textbf{ReSSL} 
& \textbf{BYOL} 
& \textbf{VICReg} 
& \textbf{NNCLR} 
& \textbf{\makecell{Barlow\\Twins}} 
& \textbf{ Ours }\\
\midrule
PathMNIST        & 92.50  & \underline{92.92} & 92.03 & 91.99 & \textbf{93.36} & 92.32 & 92.72 & 92.43   &82.59\\
BloodMNIST       & \underline{95.72} & \textbf{96.52} & 82.85 & 95.49 & 93.72 & 93.25 & 95.05  & 64.44  &91.75\\
OCTMNIST         & \textbf{79.96} & 68.82 & 60.70 & 78.56 & 74.94 & 75.20 & \underline{79.16} & 77.28    &72.00\\
BreastMNIST      & \textbf{85.26} & 82.44 & 70.38 & 72.44 & 73.21 & 73.46 & 73.08  & 79.62   &\underline{84.61}\\
PneumoniaMNIST   & 87.34 & 88.81 & 87.24 & 87.21 & \underline{90.99} & \textbf{91.03} & 88.24 & 88.08    &89.90\\
OrganAMNIST      & \underline{92.63} & 89.87 & 91.06 & 91.89 & 86.58 & 89.67 & 92.56 & 91.21    &\textbf{93.57}\\
OrganCMNIST      & 88.30 & \underline{90.22} & 84.27 & 89.31 & 86.51 & 84.88 & 89.45 & 89.29    &\textbf{91.05}\\
OrganSMNIST      & \underline{77.92} & 76.78 & 71.01 & 77.20 & 76.24 & 73.86 & 76.54 & 76.54    &\textbf{78.57}\\
RetinaMNIST      & 47.75 & 46.45 & 46.20 & 43.15 & 41.20 & 49.15 & 47.40 & \underline{52.50}    &\textbf{56.75}\\
TissueMNIST      & \underline{59.68} & 58.15 & 59.39 & 48.45 & 55.59 & 56.20 & 59.39 & 53.89    &\textbf{61.56}\\
DermaMNIST       & 73.45 & 74.22 & 68.78 & \underline{74.75} & 68.60 & 73.06 & 72.68 & 72.83    &\textbf{77.85}\\
\bottomrule
\end{tabular}
\end{table*}

\begin{table*}[t]
\centering
\scriptsize
\setlength{\tabcolsep}{1.3pt}
\renewcommand{\arraystretch}{1.15} 
\caption{Area under curve(AUC) scores of random initialization for self-supervised training using ResNet-50 across different methods and datasets.}
\label{MIMVcomparision2}
\begin{tabular}{l|c|c|c|c|c|c|c|c|c}
\toprule
\textbf{Dataset}
& \textbf{MoCoV3} 
& \textbf{SimCLR}
& \textbf{DINO} 
& \textbf{ReSSL} 
& \textbf{BYOL} 
& \textbf{VICReg} 
& \textbf{NNCLR} 
& \textbf{\makecell{Barlow\\Twins}} 
& \textbf{ Ours }\\
\midrule
PathMNIST        &99.27  &\textbf{99.45}  &99.28  &\underline{99.23}  &99.44  & 99.25 &99.26  &99.38    &97.96\\
BloodMNIST       &99.71  & \textbf{99.80} &97.53  &99.69  &99.47  & 99.47 & \underline{99.74} &90.40    &99.28\\
OCTMNIST         &96.81  &92.17  &90.01  &96.07  &95.67  &96.60  &\textbf{97.11}  &\underline{97.05}    &96.59\\
BreastMNIST      &80.93  &\underline{85.56}  &62.23  & 51.65 & 53.18 & 58.48 & 63.59 & 67.44   &\textbf{88.37}\\
PneumoniaMNIST   &\textbf{98.39}  &94.36  &90.29  & 94.23 &97.45  &96.50  &94.27  &92.38    &\underline{98.19}\\
OrganAMNIST      &99.57  &97.92  &90.49  &99.60  &98.96  &99.21  &\underline{99.67}  &99.47    &\textbf{99.73}\\
OrganCMNIST      &98.51  &99.29  & 98.35 &99.30  &98.71  &98.67  &\underline{99.31}  &98.90    &\textbf{99.40}\\
OrganSMNIST      &\textbf{97.87}  &97.35  &95.55  &97.57  &97.41  &96.47  &97.40  &97.57    &\underline{97.70}\\
RetinaMNIST      &65.37  &63.38  &58.72  &60.35  &51.64  &63.49  &62.72  & \underline{67.19}   &\textbf{76.02}\\
TissueMNIST      &\underline{88.66}  &87.38  &88.55  &78.65  &86.11  &86.34  &88.41  &84.16    &\textbf{89.77}\\
DermaMNIST       &87.60  &89.20  &80.67  &\underline{89.87}  &76.65  & 86.64 & 85.44 & 86.28   &\textbf{93.60}\\
\bottomrule
\end{tabular}
\end{table*}

\begin{figure*}[h!]
    \centering

    \subfloat[PathMNIST]{%
        \includegraphics[width=0.32\linewidth]{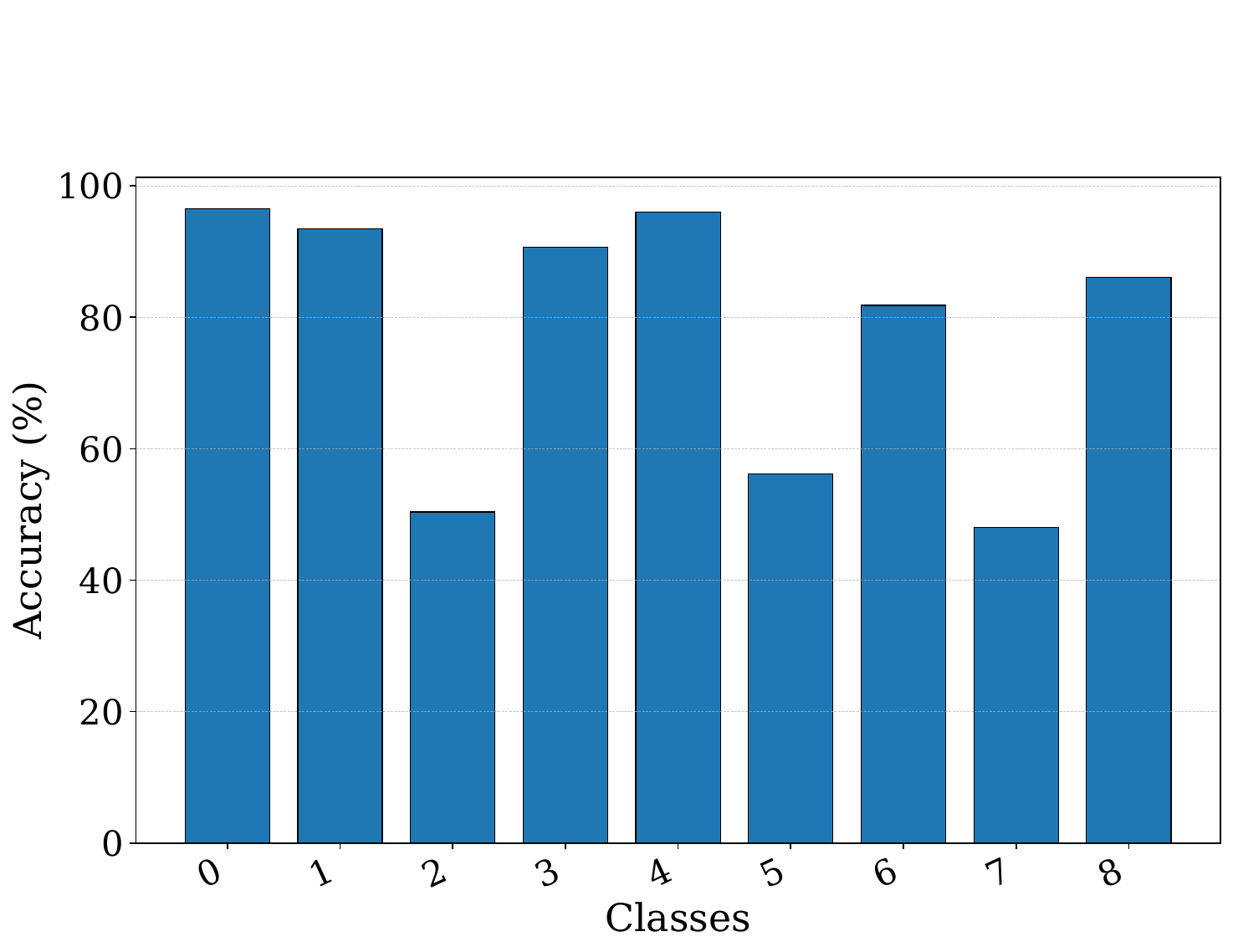}
    }\hfill
    \subfloat[BloodMNIST]{%
        \includegraphics[width=0.32\linewidth]{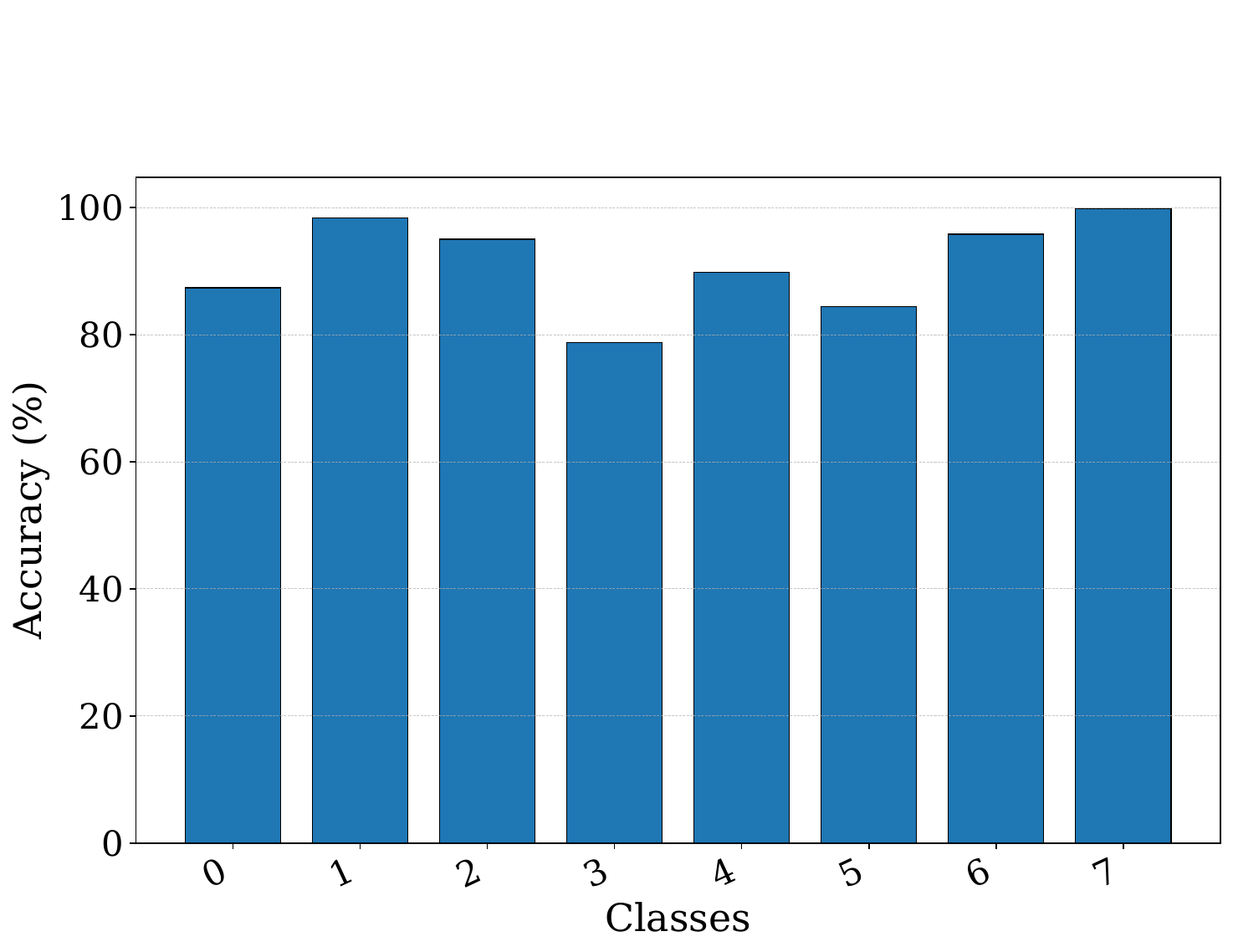}
    }\hfill
    \subfloat[OCTMNIST]{%
        \includegraphics[width=0.32\linewidth]{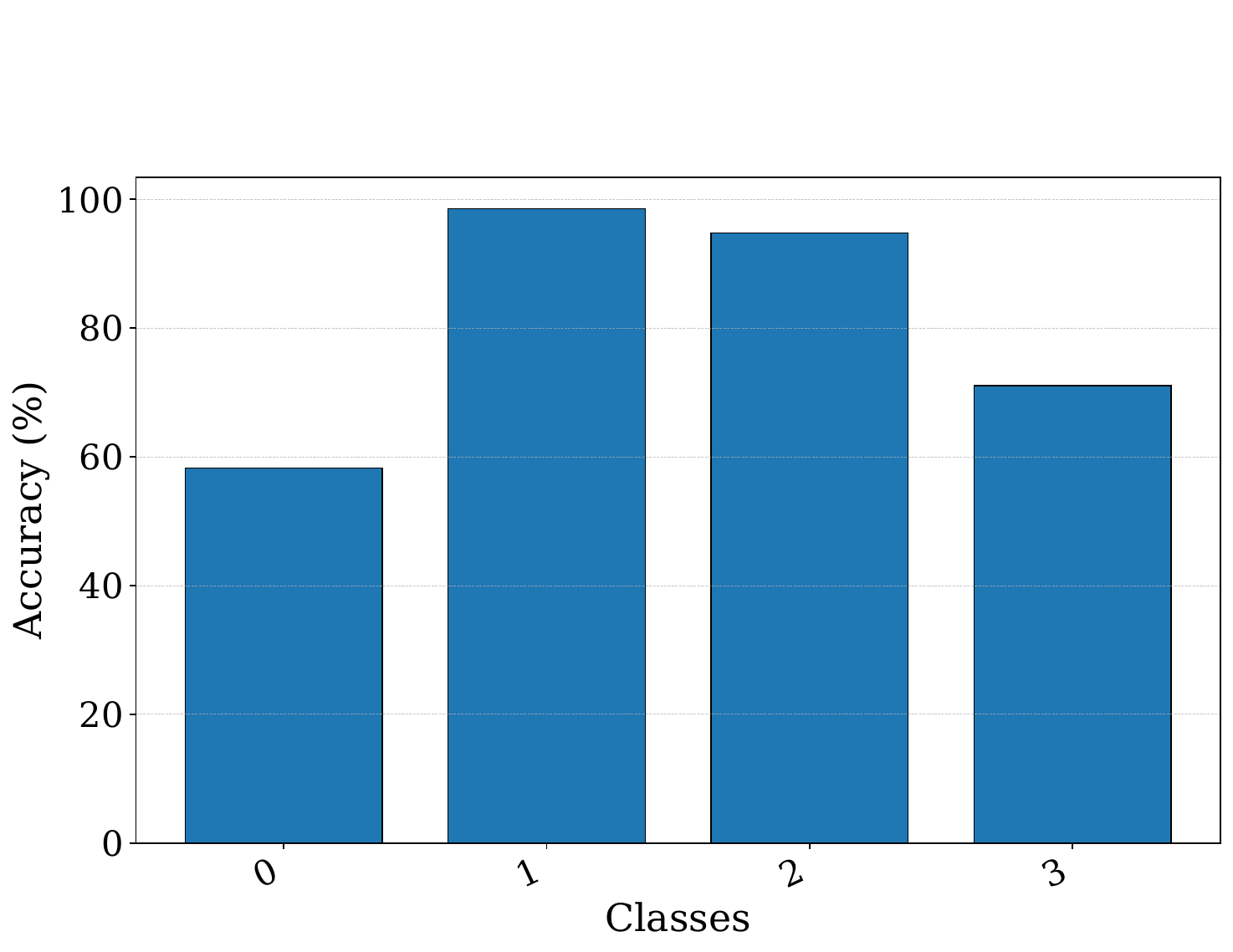}
    }

    \vspace{0.25cm}

    \subfloat[BreastMNIST]{%
        \includegraphics[width=0.32\linewidth]{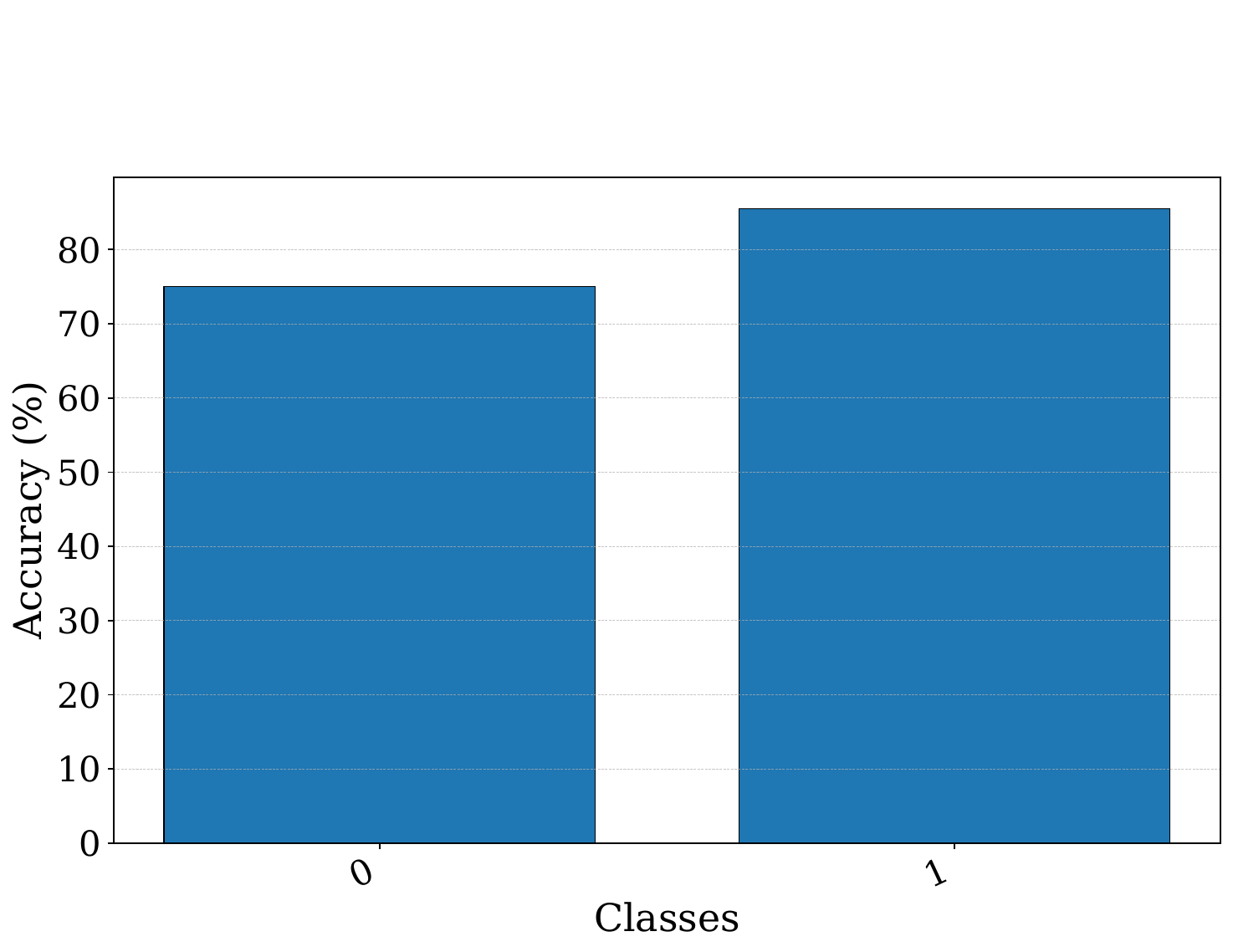}
    }\hfill
    \subfloat[PneumoniaMNIST]{%
        \includegraphics[width=0.32\linewidth]{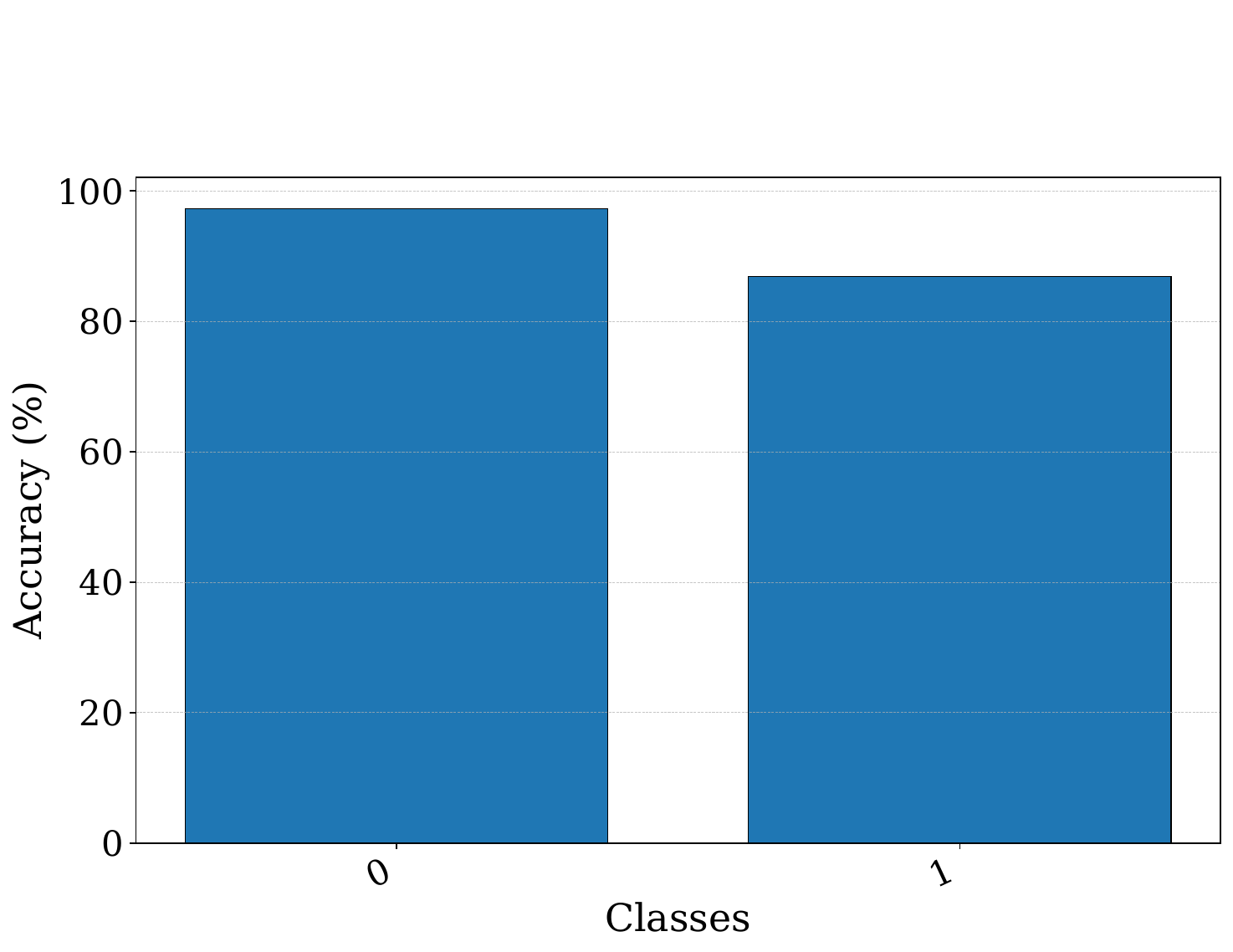}
    }\hfill
    \subfloat[OrganAMNIST]{%
        \includegraphics[width=0.32\linewidth]{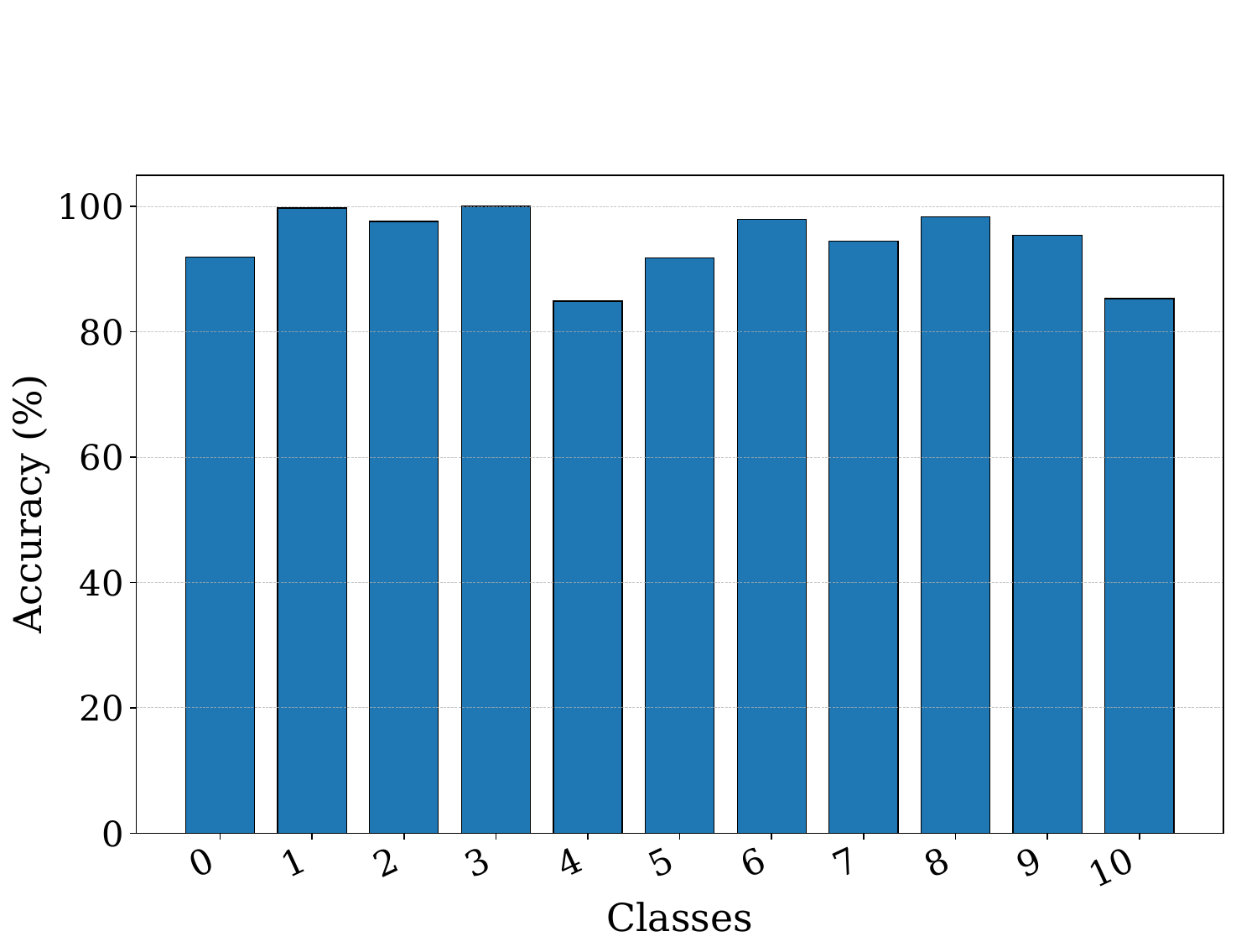}
    }

    \vspace{0.25cm}

    \subfloat[OrganCMNIST]{%
        \includegraphics[width=0.32\linewidth]{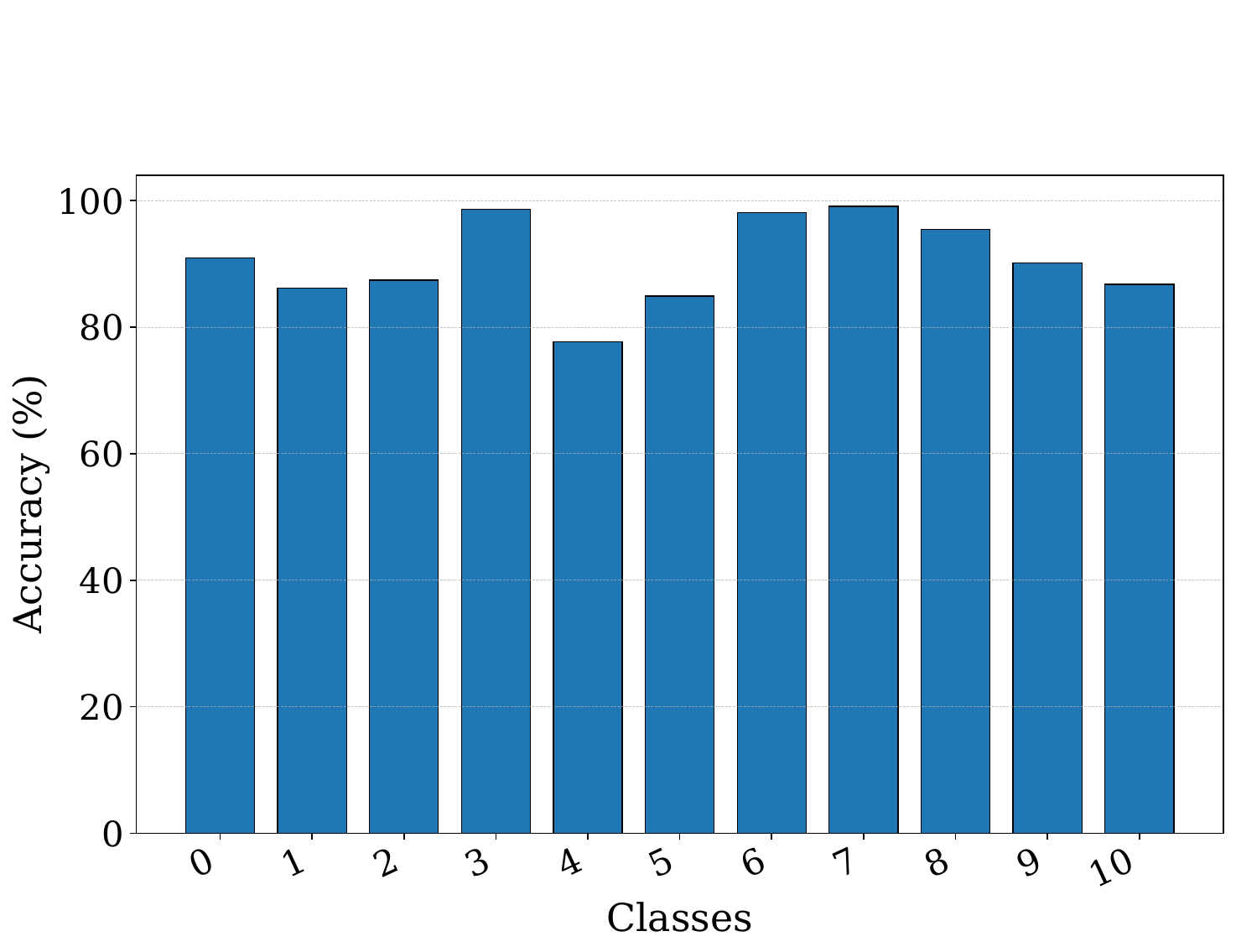}
    }\hfill
    \subfloat[OrganSMNIST]{%
        \includegraphics[width=0.32\linewidth]{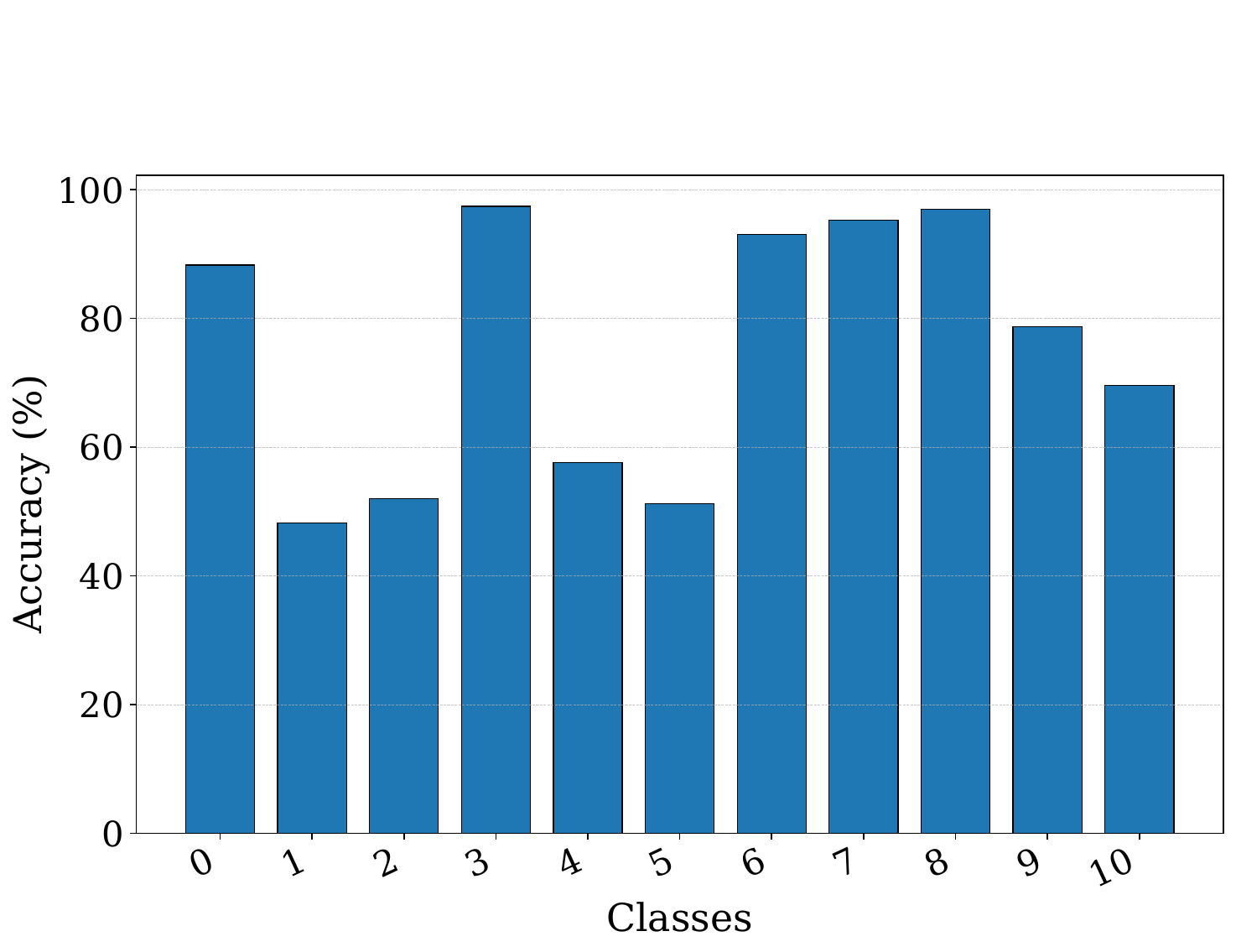}
    }\hfill
    \subfloat[RetinaMNIST]{%
        \includegraphics[width=0.32\linewidth]{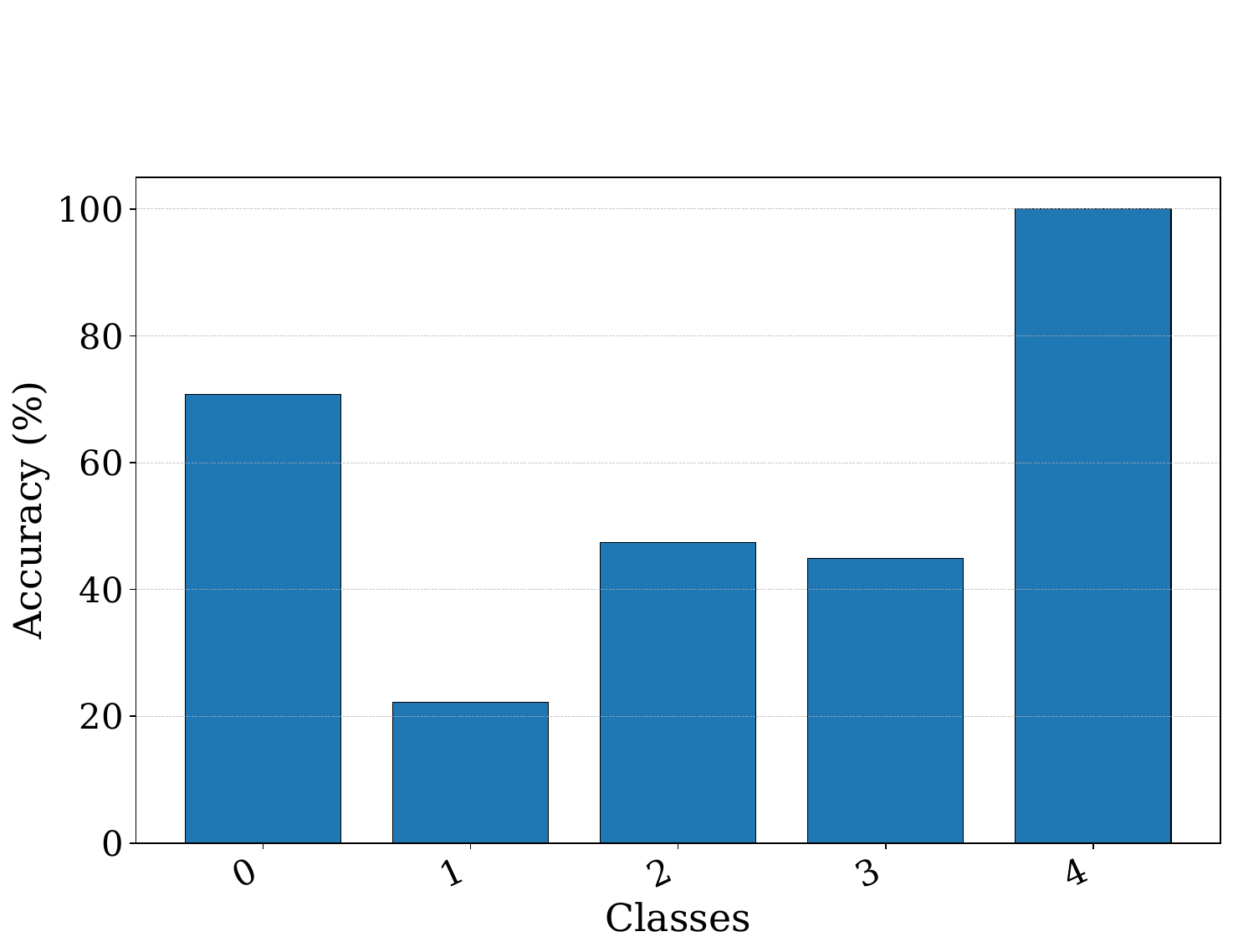}
    }

    \vspace{0.25cm}

    \subfloat[TissueMNIST]{%
        \includegraphics[width=0.32\linewidth]{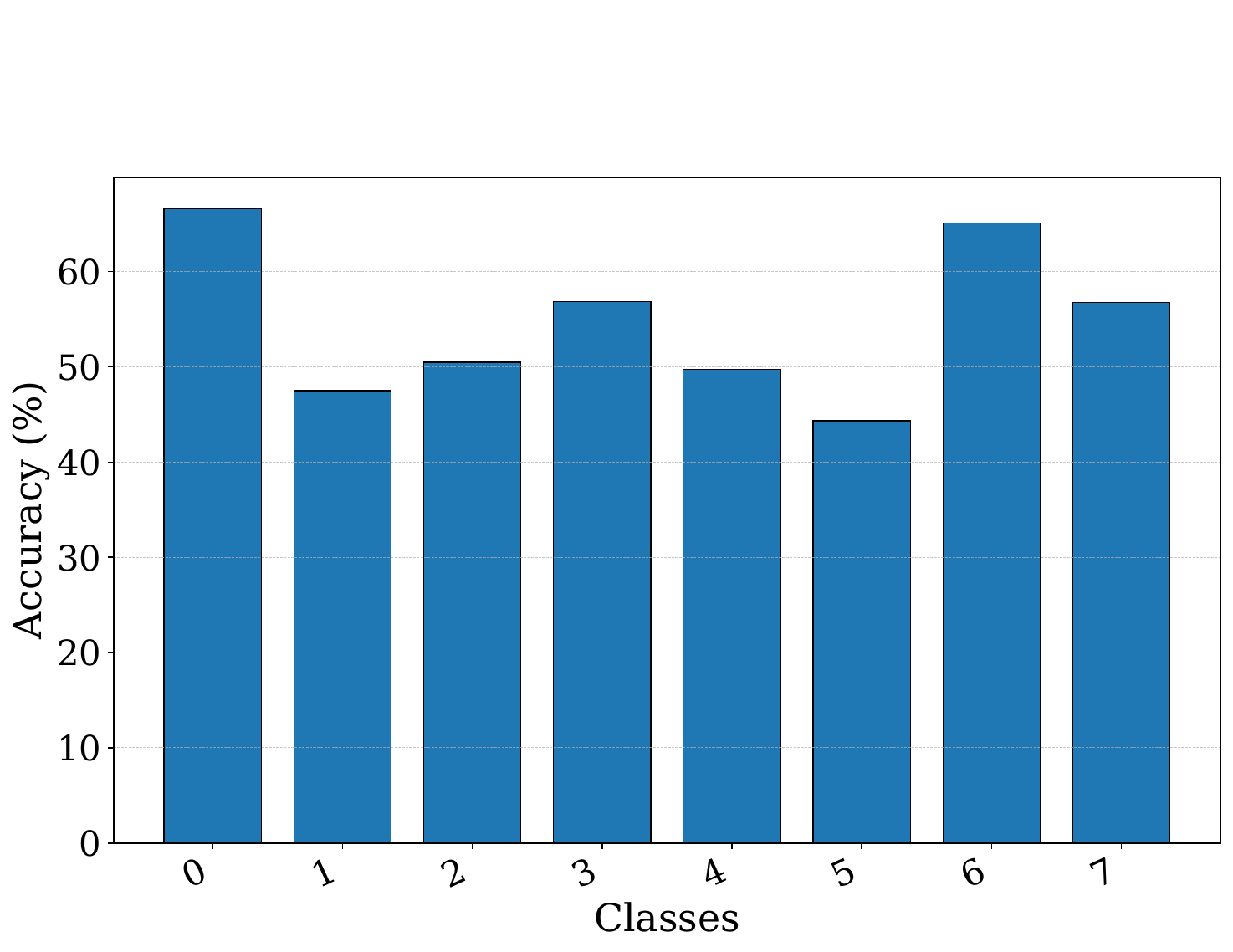}
    }\hfill
    \subfloat[DermaMNIST]{%
        \includegraphics[width=0.32\linewidth]{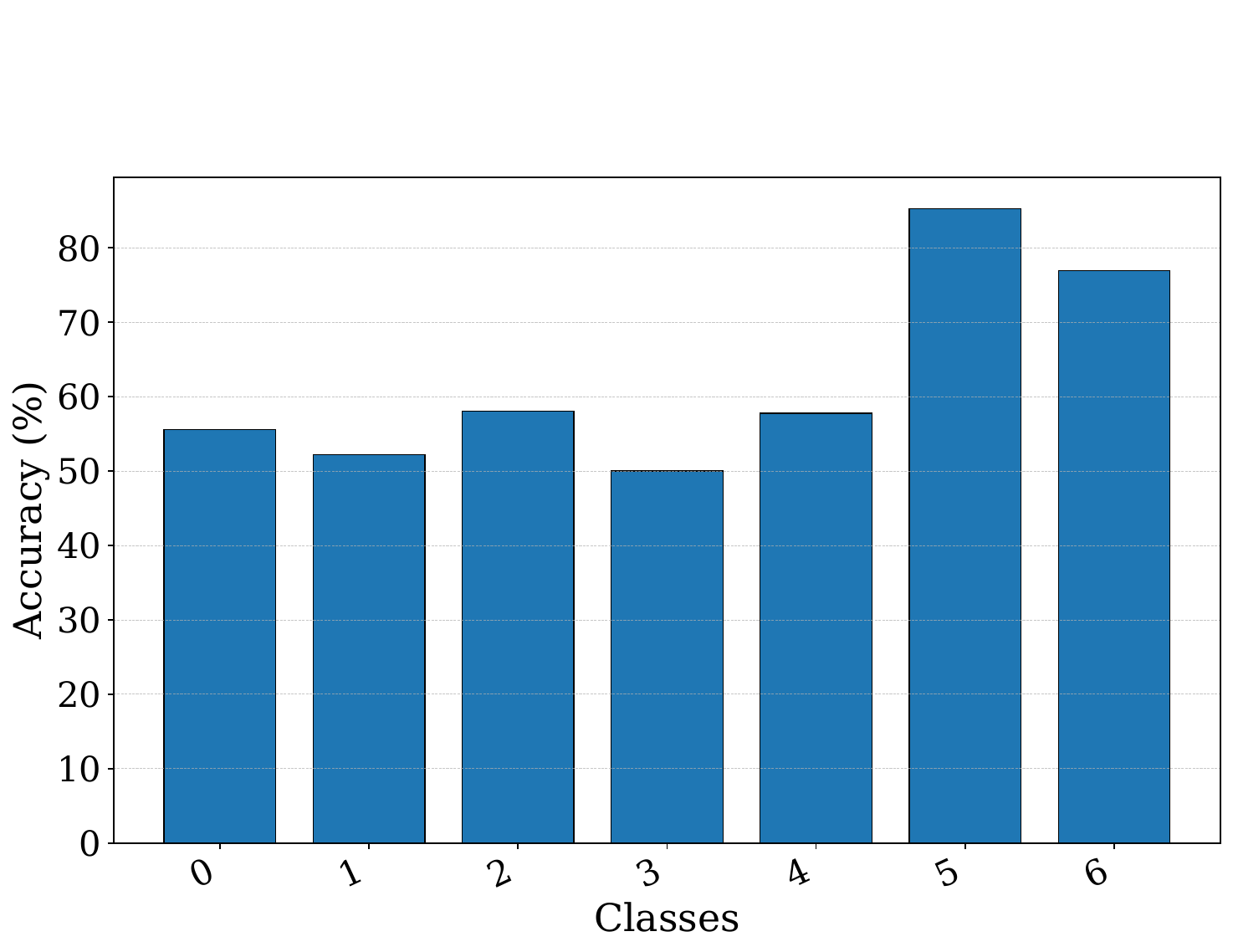}
    }

    \caption{Per-class accuracy of AMIMV-SSL on MedMNIST datasets.}
    \label{fig:perclass}
\end{figure*}

\begin{figure*}[h!]
    \centering

    \subfloat[PathMNIST]{%
        \includegraphics[width=0.32\linewidth]{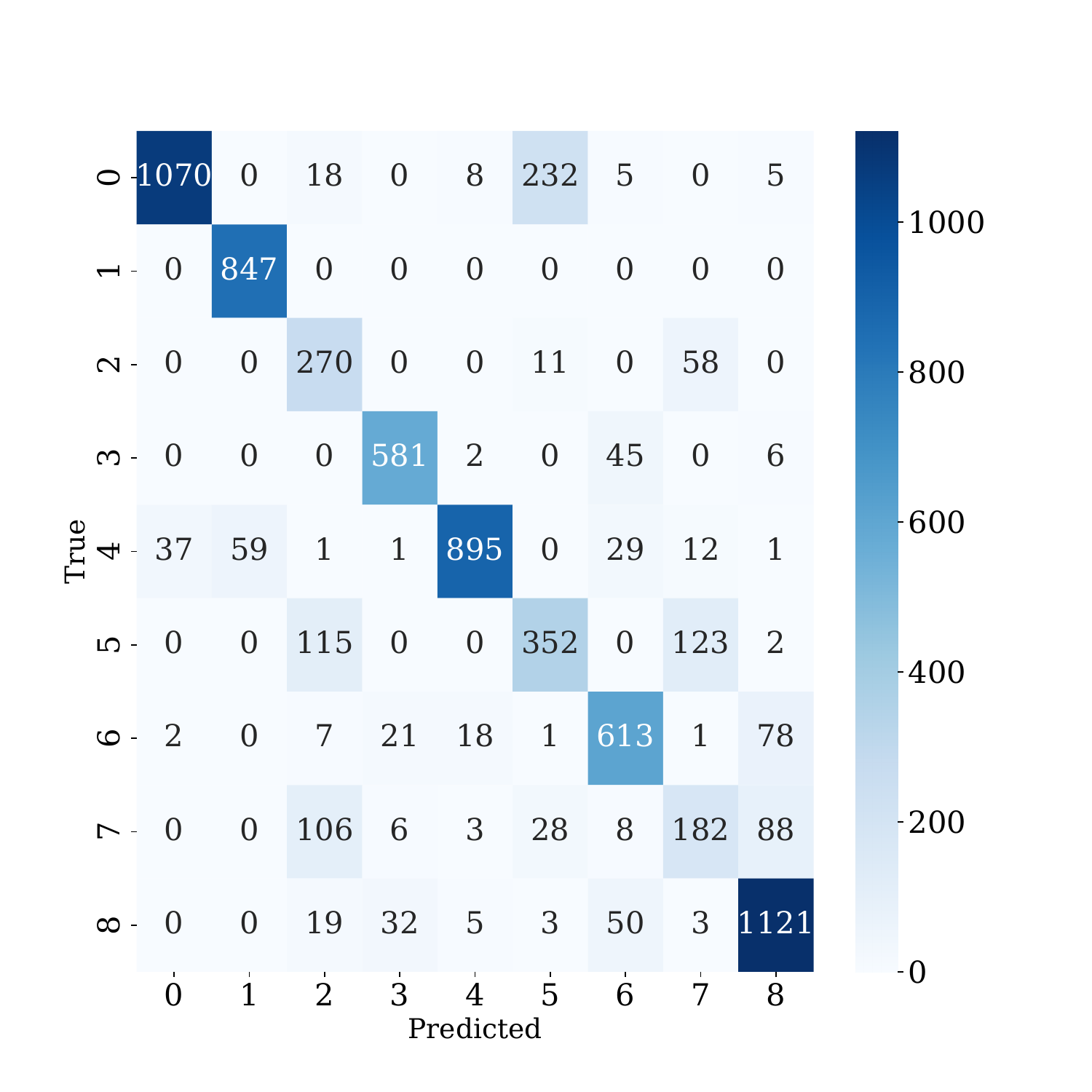}
    }\hfill
    \subfloat[BloodMNIST]{%
        \includegraphics[width=0.32\linewidth]{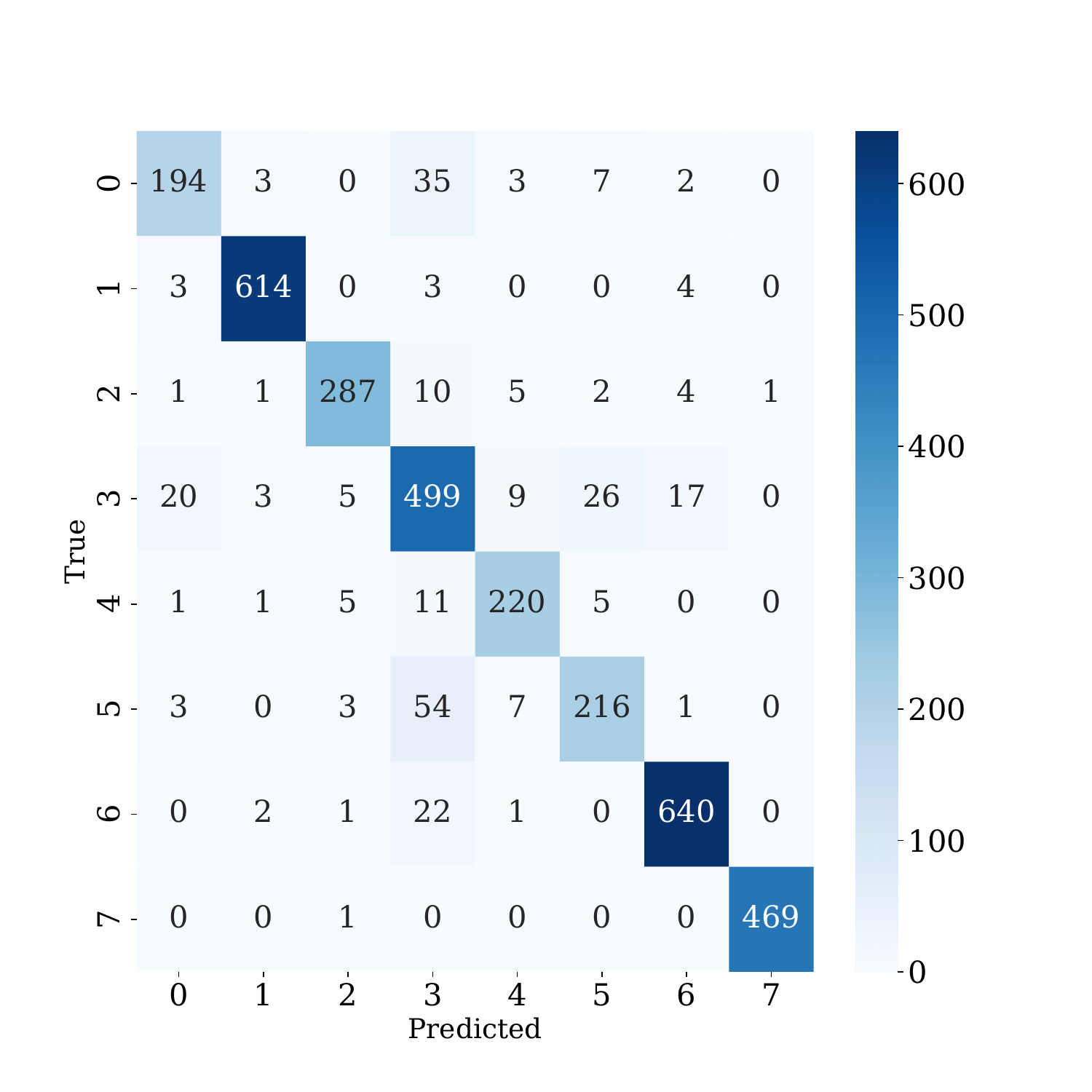}
    }\hfill
    \subfloat[OCTMNIST]{%
        \includegraphics[width=0.32\linewidth]{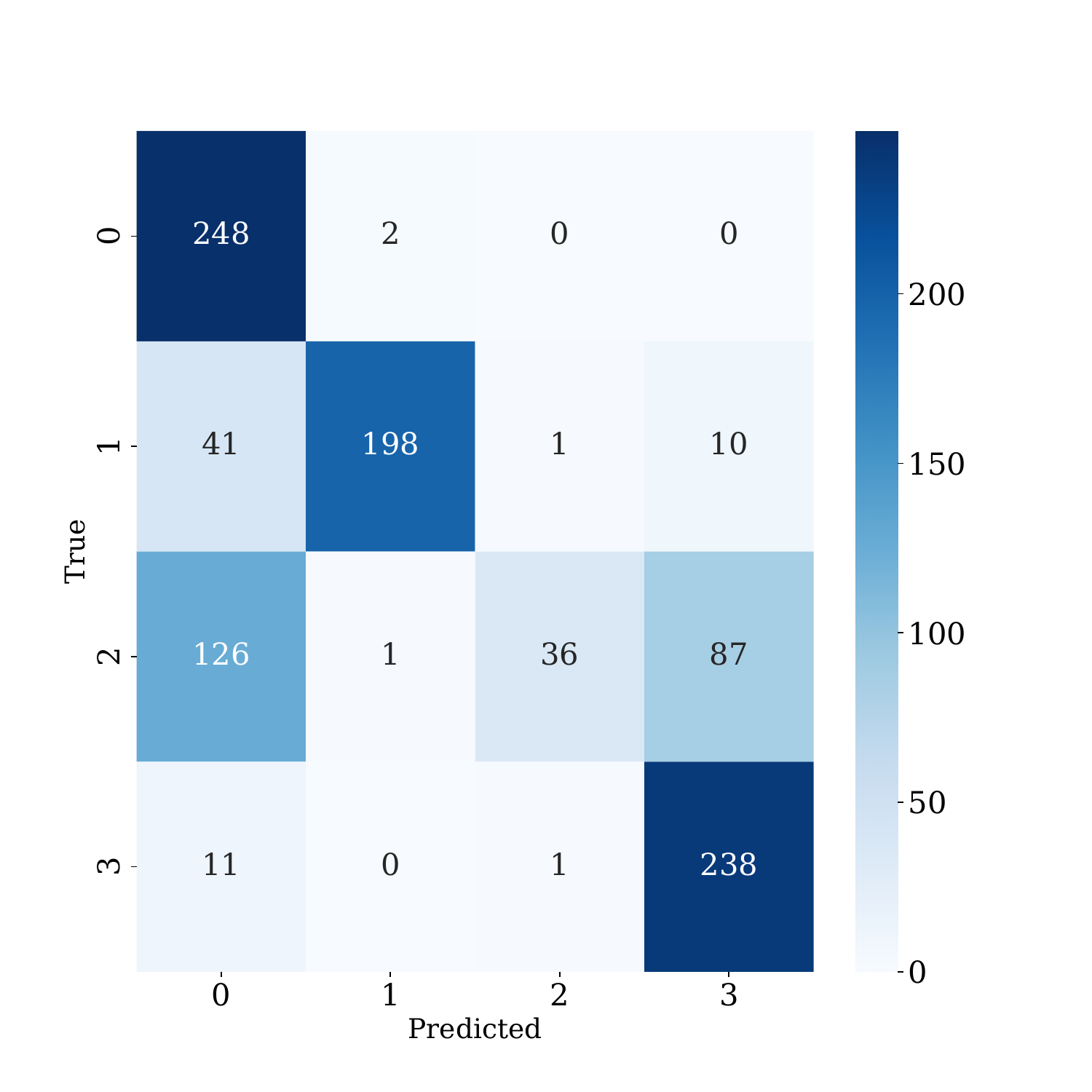}
    }
    \vspace{0.25cm}
    \subfloat[BreastMNIST]{%
        \includegraphics[width=0.32\linewidth]{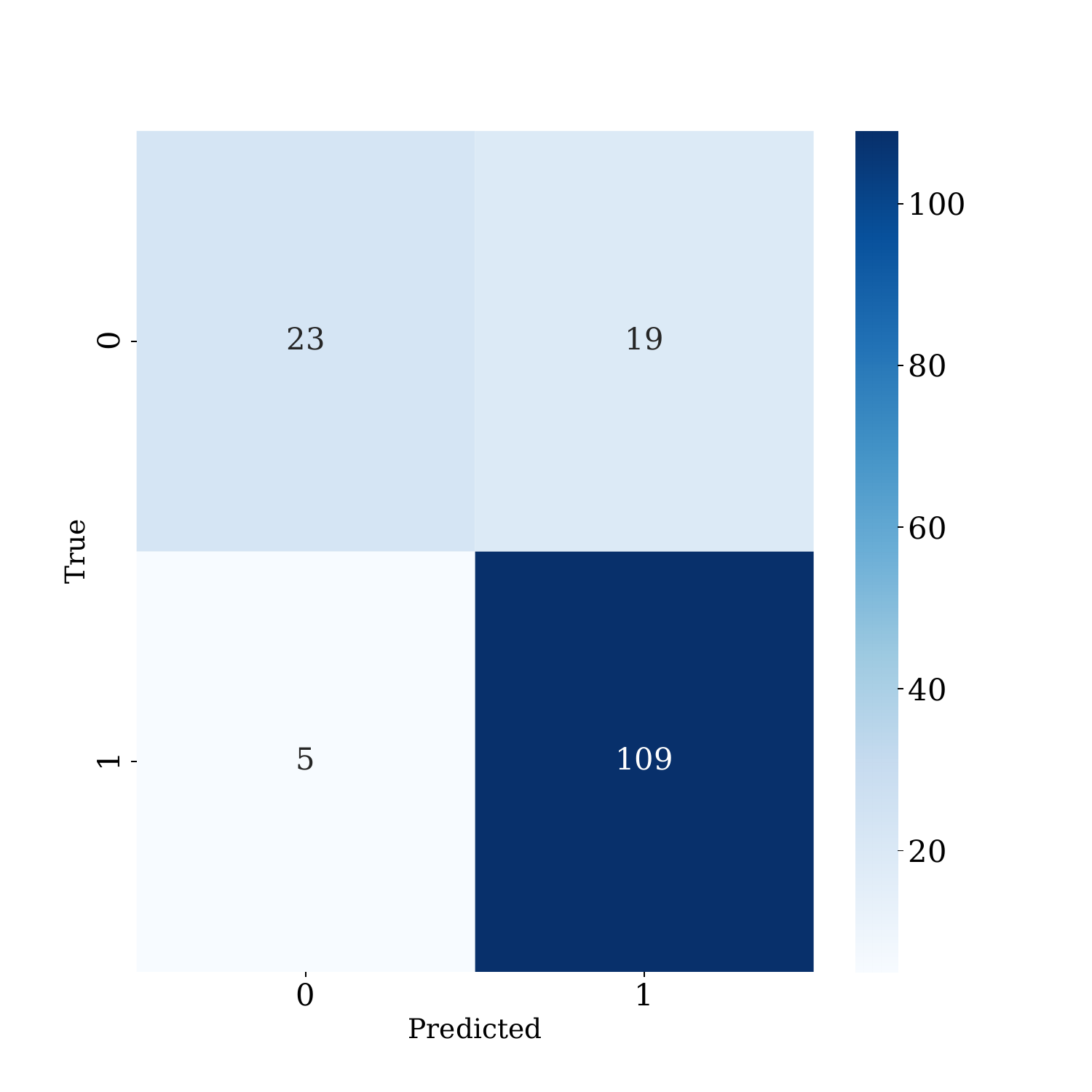}
    }\hfill
    \subfloat[PneumoniaMNIST]{%
        \includegraphics[width=0.32\linewidth]{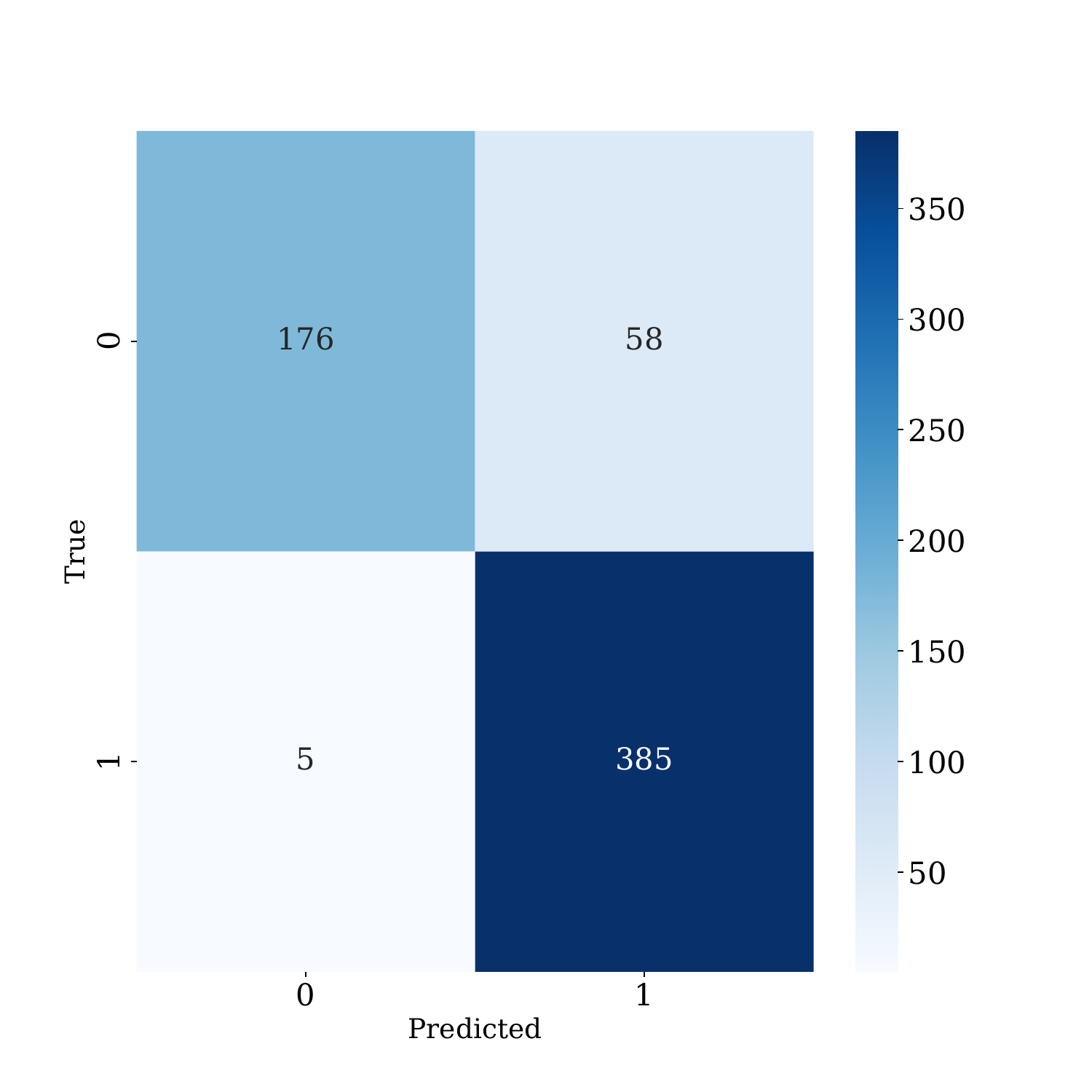}
    }\hfill
    \subfloat[OrganAMNIST]{%
        \includegraphics[width=0.32\linewidth]{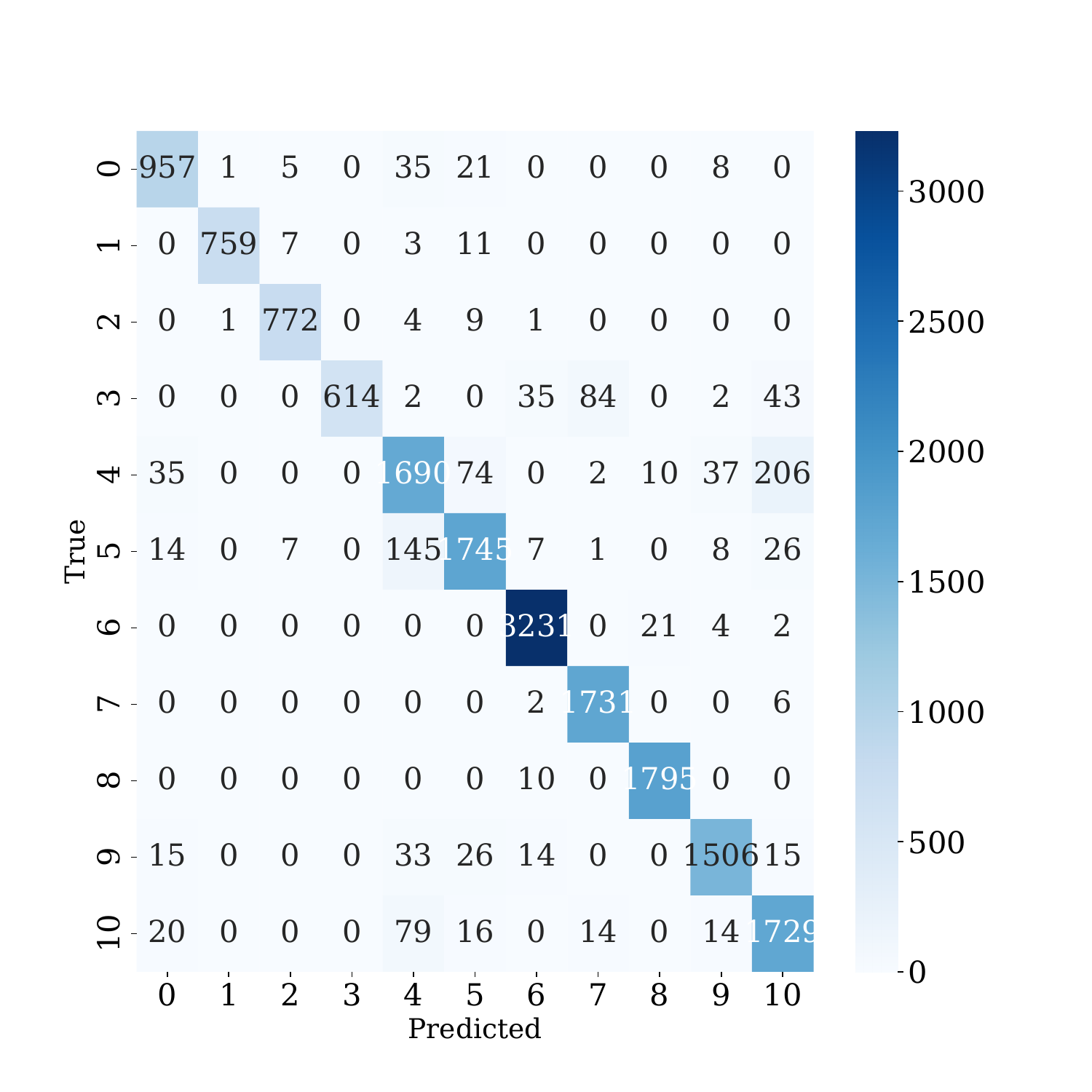}
    }

    \vspace{0.25cm}

    \subfloat[OrganCMNIST]{%
        \includegraphics[width=0.32\linewidth]{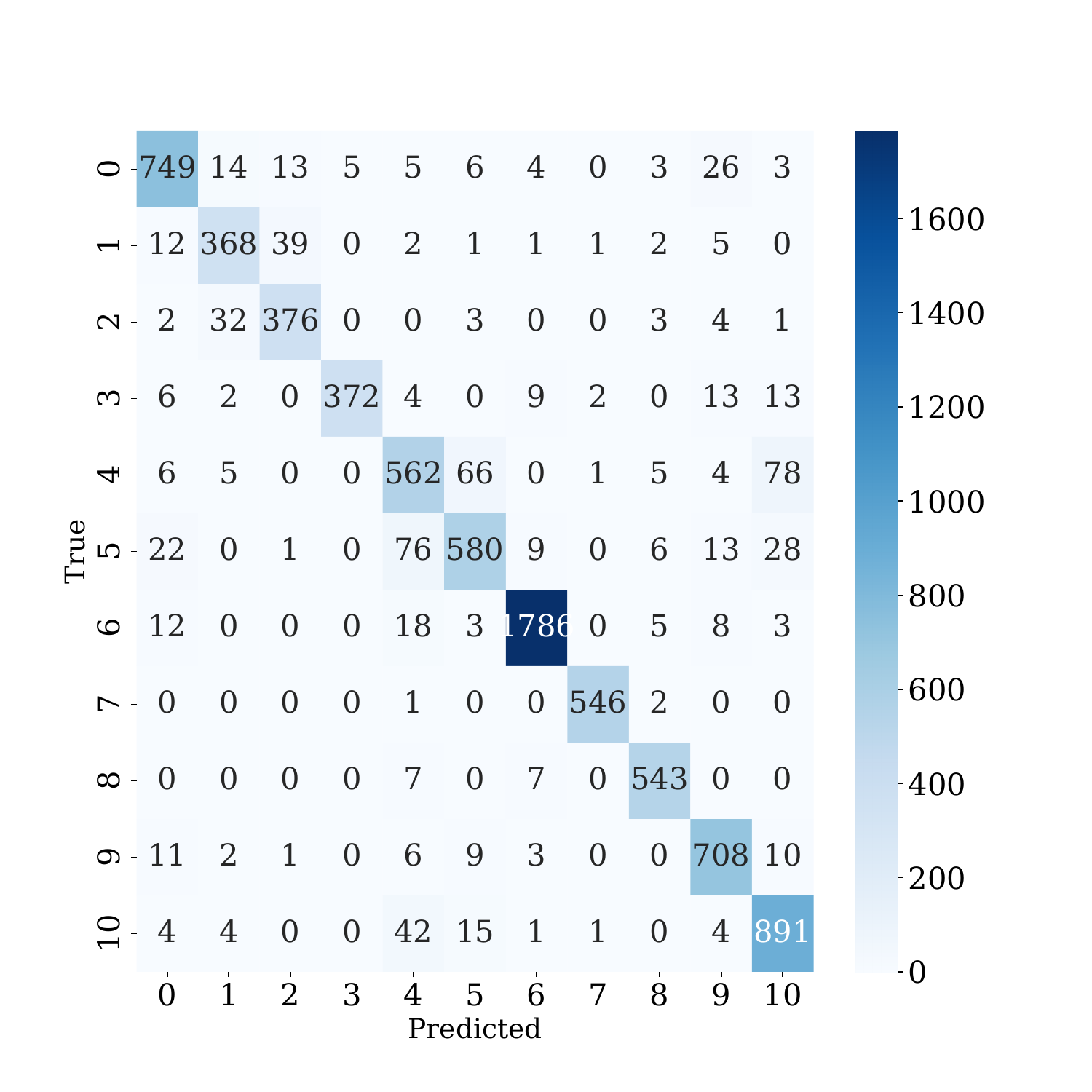}
    }\hfill
    \subfloat[OrganSMNIST]{%
        \includegraphics[width=0.32\linewidth]{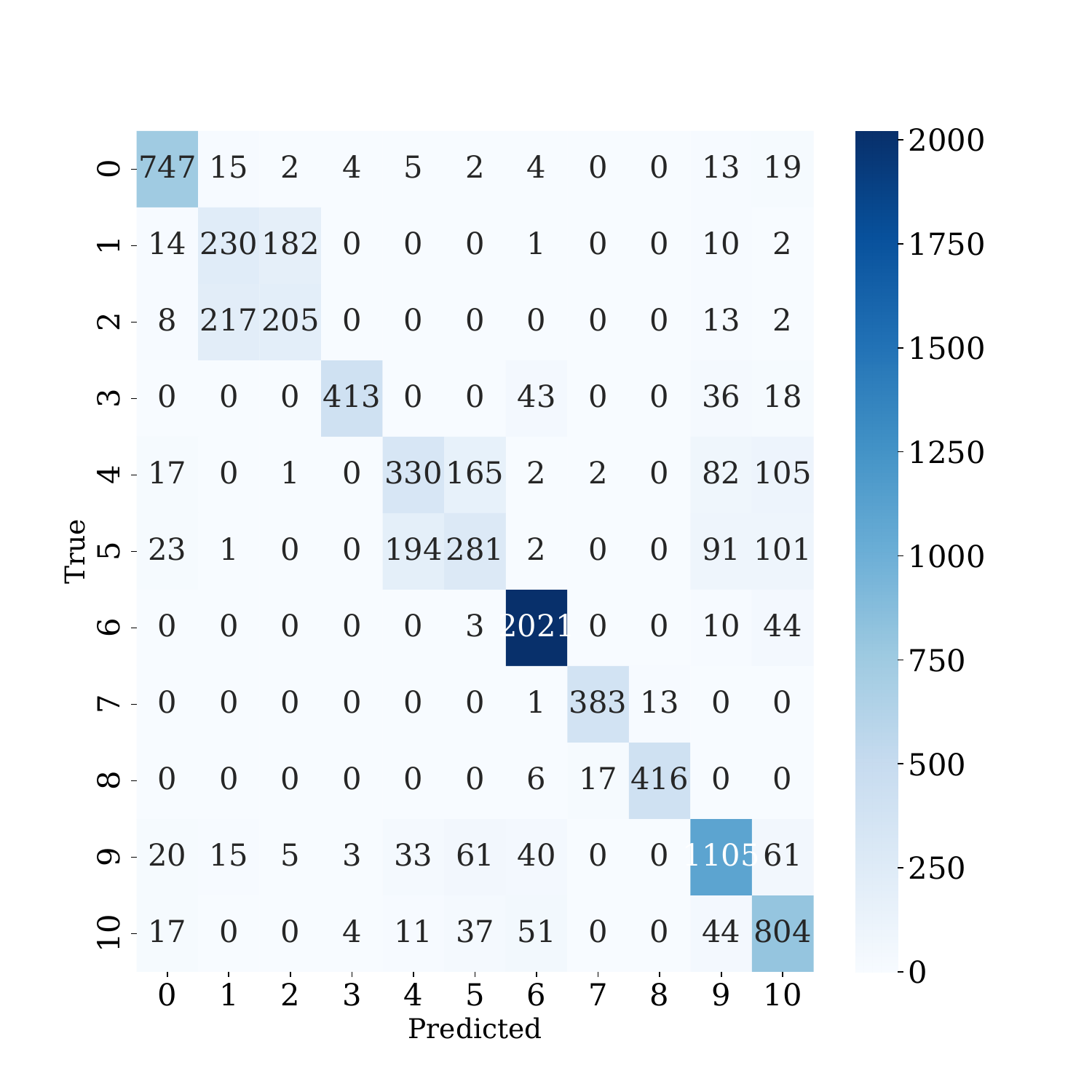}
    }\hfill
    \subfloat[RetinaMNIST]{%
        \includegraphics[width=0.32\linewidth]{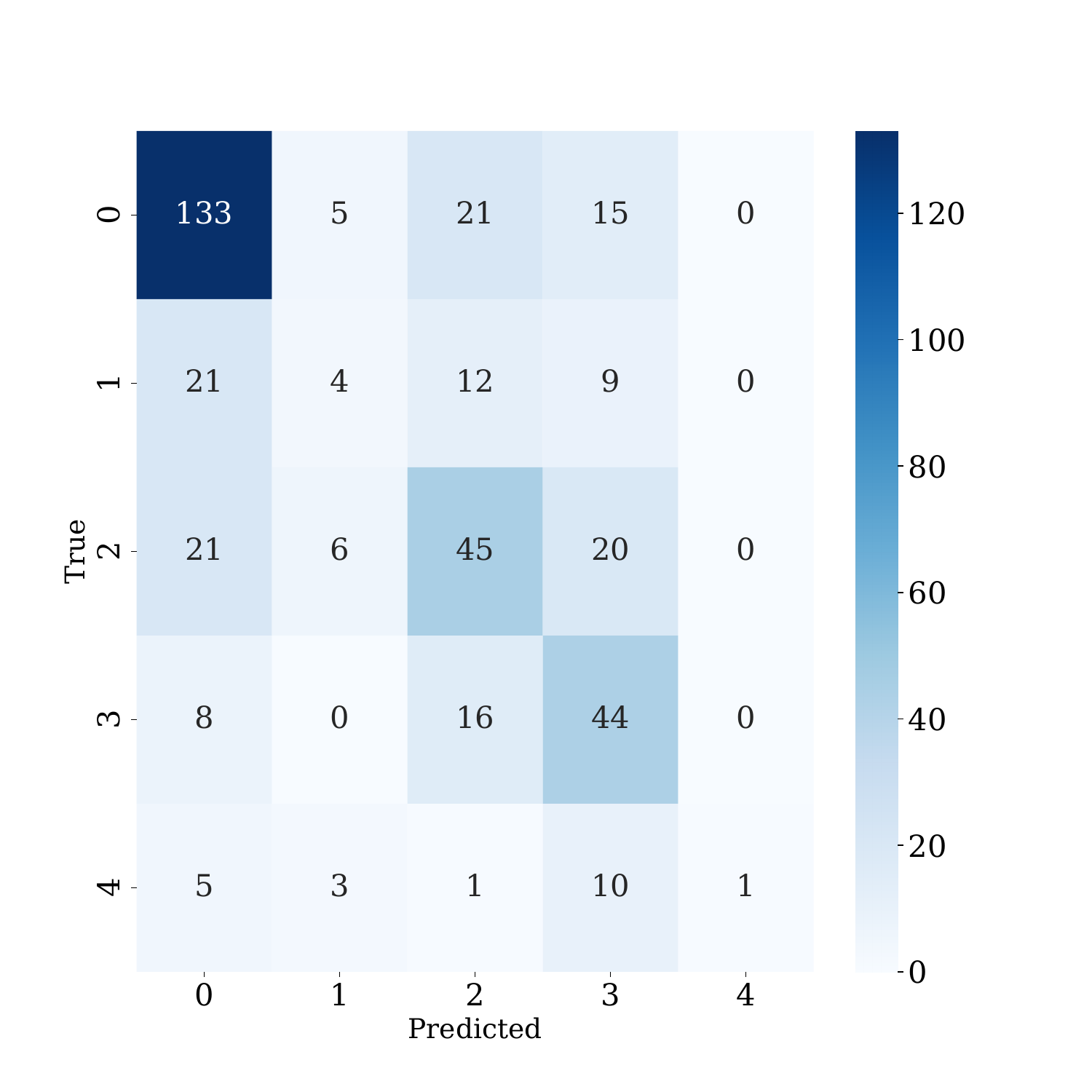}
    }

    \vspace{0.25cm}

    \subfloat[TissueMNIST]{%
        \includegraphics[width=0.32\linewidth, height=0.3\linewidth]{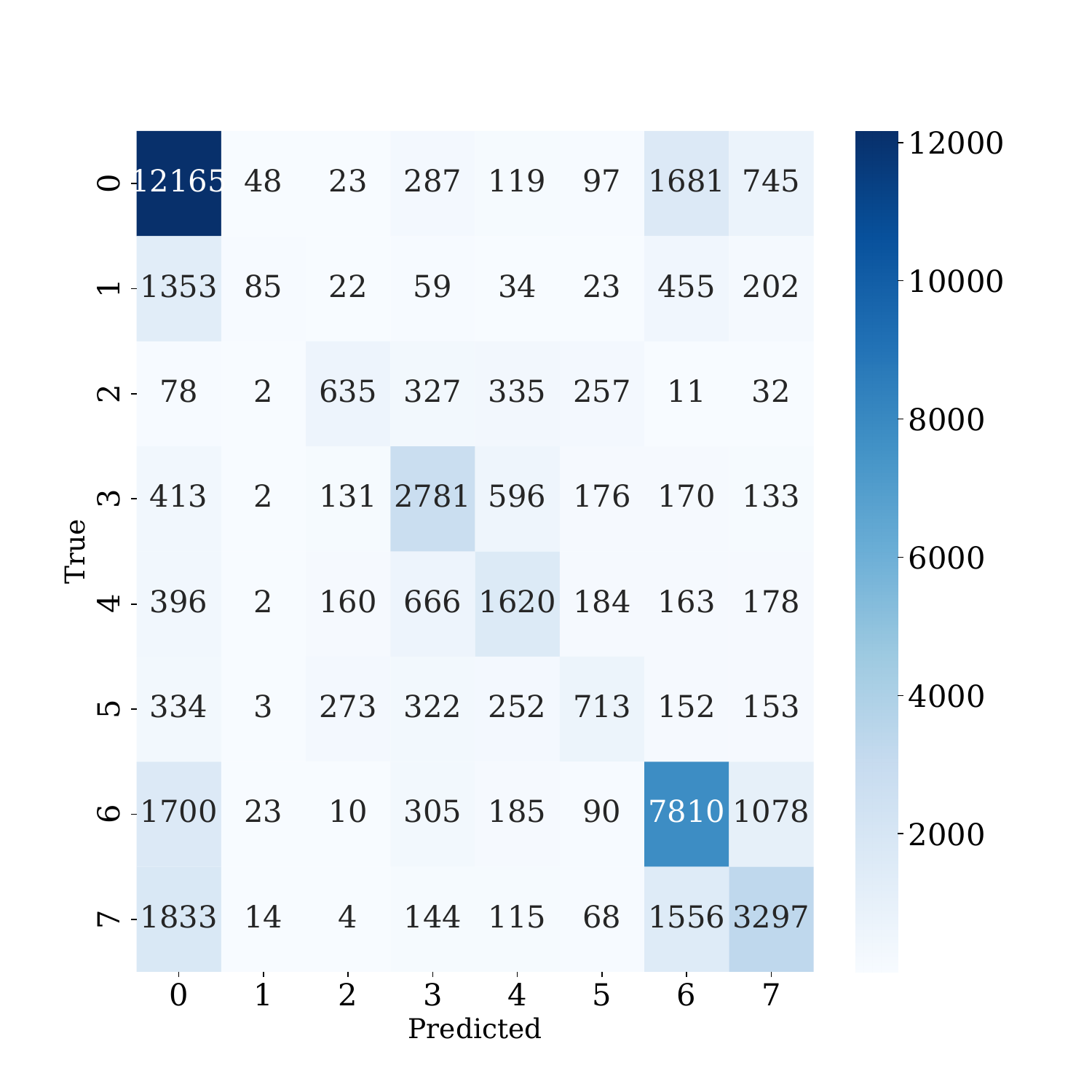}
    }\hfill
    \subfloat[DermaMNIST]{%
        \includegraphics[width=0.32\linewidth, height=0.3\linewidth]{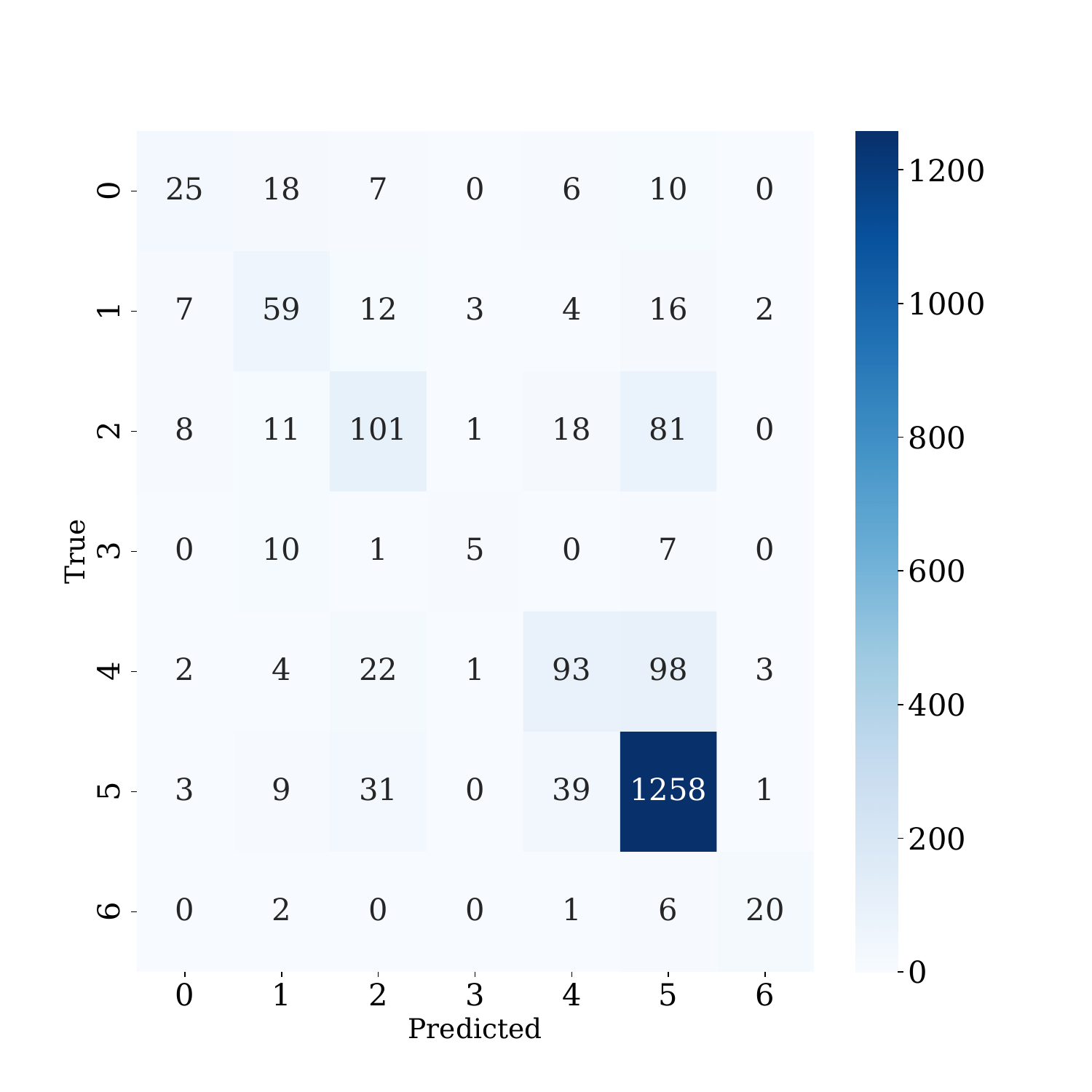}
    }
    \caption{Confusion matrices of AMIMV-SSL across eleven MedMNIST datasets, including BloodMNIST, BreastMNIST, DermaMNIST, OCTMNIST, PathMNIST, PneumoniaMNIST, RetinaMNIST, TissueMNIST, OrganAMNIST, and OrganCMNIST}
    \label{fig:confusion}
\end{figure*}

Table.\ref{MIMVcomparision1}and Table.\ref{MIMVcomparision2} demonstrate that our method consistently surpasses existing SSL techniques on both partially imbalanced and imbalanced MedMNIST datasets. The most significant improvements are observed on RetinaMNIST, TissueMNIST, and DermaMNIST, which have high imbalance ratios and limited representation of minority classes. Throughout the various OrganMNIST datasets, our method shows consistent enhancements, signifying its resilience to moderate distribution skew. In contrast, for fairly balanced datasets like PathMNIST, BloodMNIST, and OCTMNIST, the performance variations between SSL techniques are minimal and statistically less significant. 
These results verify that our proposed method performs particularly well in the presence of a class imbalance.
Another crucial observation that can be made from Table.\ref{MIMVcomparision1} is that every SSL model does not perform well on all datasets. For instance, MocoV3 does not perform well on RetinaMNIST, SIMCLR on RetinaMNIST, Dino on RetinaMNIST/BreastMNIST and so on. The same pattern can be observed from Table.\ref{MIMVcomparision2}. In contrast, AMIMV is consistent across the entire range of datasets, whether it is balanced or imbalanced. In other words, this means that AMIMV learning never drops drastically. 
Figure~\ref{fig:perclass} shows the per-class accuracy of AMIMV-SSL across all MedMNIST datasets. Despite severe class imbalance and data set-specific imbalance, AMIMV-SSL maintains stable performance across frequent and rare classes. In particular, minority classes consistently achieve higher accuracy compared to conventional self-supervised baselines, indicating reduced bias toward dominant classes. Figure~\ref{fig:confusion} shows the confusion matrices of AMIMV-SSL in eleven MedMNIST datasets. Across all datasets, AMIMV-SSL exhibits strong diagonal dominance, indicating reliable class-wise discrimination despite severe class imbalance. Importantly, rare and minority classes demonstrate improved recall with reduced confusion toward dominant classes, highlighting the effectiveness of asymmetric multi-image, multi-view representation learning. These qualitative results align with the quantitative performance gains and confirm that AMIMV-SSL mitigates head-class domination in real-world medical datasets. 
\FloatBarrier
\section{Conclusions and Future Work}
The main contribution of this work is in addressing the dataset scarcity as well as dataset imbalance problem related to medical image datasets. To achieve this end, we made use of our self supervised learning framework~\cite{sharma2025maximally} in which a novel augmentation strategy was introduced. By adapting this extended framework in the domain of medical image classification, we demonstrated that the asymmetric multi-image multi-view is more robust than the traditional multi-view SSL frameworks for addressing higher degrees of imbalance in medical image datasets.
Through extensive evaluation on the MedMNIST benchmark, covering eleven diverse medical imaging datasets, we demonstrated that AMIMV-SSL consistently improves representation robustness, minority-class performance, and generalization under long-tailed distributions.
In future work, we plan to extend AMIMV-SSL to high-resolution medical imaging datasets, as well as to 3D and multi-modal imaging scenarios. We also aim to investigate the applicability of the proposed framework to vision–language tasks, including medical visual question answering (VQA).

%
%
%
%
%

\bibliographystyle{apalike}
\bibliography{references}

@article{azizi23nature,
  title={Robust and data-efficient generalization of self-supervised machine learning for diagnostic imaging},
  author={Azizi, Shekoofeh and Laura, Culp and Jan, Freyberg and ... and Vivek, Natarajan},
  journal={Nature Biomedical Engineering},
  volume={7},
  number={6},
pages={756-779},
  year={2023}
}

@article{AljuaidA22survey,
  author       = {Abeer Aljuaid and
                  Mohd Anwar},
  title        = {Survey of Supervised Learning for Medical Image Processing},
  journal      = {{SN} Comput. Sci.},
  volume       = {3},
  number       = {4},
  pages        = {292},
  year         = {2022},
}

@inproceedings{caron2021emerging,
  title={Emerging properties in self-supervised vision transformers},
  author={Caron, Mathilde and Touvron, Hugo and Misra, Ishan and J{\'e}gou, Herv{\'e} and Mairal, Julien and Bojanowski, Piotr and Joulin, Armand},
  booktitle={Proceedings of the IEEE/CVF international conference on computer vision},
  pages={9650--9660},
  year={2021}
}

@article{zheng2021ressl,
  title={Ressl: Relational self-supervised learning with weak augmentation},
  author={Zheng, Mingkai and You, Shan and Wang, Fei and Qian, Chen and Zhang, Changshui and Wang, Xiaogang and Xu, Chang},
  journal={Advances in Neural Information Processing Systems},
  volume={34},
  pages={2543--2555},
  year={2021}
}

@inproceedings{chen2021empirical,
  title={An empirical study of training self-supervised vision transformers},
  author={Chen, Xinlei and Xie, Saining and He, Kaiming},
  booktitle={Proceedings of the IEEE/CVF international conference on computer vision},
  pages={9640--9649},
  year={2021}
}

@inproceedings{dwibedi2021little,
  title={With a little help from my friends: Nearest-neighbor contrastive learning of visual representations},
  author={Dwibedi, Debidatta and Aytar, Yusuf and Tompson, Jonathan and Sermanet, Pierre and Zisserman, Andrew},
  booktitle={Proceedings of the IEEE/CVF international conference on computer vision},
  pages={9588--9597},
  year={2021}
}

@article{bardes2021vicreg,
  title={Vicreg: Variance-invariance-covariance regularization for self-supervised learning},
  author={Bardes, Adrien and Ponce, Jean and LeCun, Yann},
  journal={arXiv preprint arXiv:2105.04906},
  year={2021}
}

@inproceedings{zbontar2021barlow,
  title={Barlow twins: Self-supervised learning via redundancy reduction},
  author={Zbontar, Jure and Jing, Li and Misra, Ishan and LeCun, Yann and Deny, St{\'e}phane},
  booktitle={International conference on machine learning},
  pages={12310--12320},
  year={2021},
  organization={PMLR}
}

@article{yang2023medmnist,
  title={Medmnist v2-a large-scale lightweight benchmark for 2d and 3d biomedical image classification},
  author={Yang, Jiancheng and Shi, Rui and Wei, Donglai and Liu, Zequan and Zhao, Lin and Ke, Bilian and Pfister, Hanspeter and Ni, Bingbing},
  journal={Scientific Data},
  volume={10},
  number={1},
  pages={41},
  year={2023},
  publisher={Nature Publishing Group UK London}
}

@inproceedings{goyal2019scaling,
  title={Scaling and benchmarking self-supervised visual representation learning},
  author={Goyal, Priya and Mahajan, Dhruv and Gupta, Abhinav and Misra, Ishan},
  booktitle={Proceedings of the ieee/cvf International Conference on computer vision},
  pages={6391--6400},
  year={2019}
}

@article{marks2025closer,
  title={A closer look at benchmarking self-supervised pre-training with image classification},
  author={Marks, Markus and Knott, Manuel and Kondapaneni, Neehar and Cole, Elijah and Defraeye, Thijs and Perez-Cruz, Fernando and Perona, Pietro},
  journal={International Journal of Computer Vision},
  pages={1--13},
  year={2025},
  publisher={Springer}
}

@article{doerrich2025rethinking,
  title={Rethinking model prototyping through the MedMNIST+ dataset collection},
  author={Doerrich, Sebastian and Di Salvo, Francesco and Brockmann, Julius and Ledig, Christian},
  journal={Scientific reports},
  volume={15},
  number={1},
  pages={7669},
  year={2025},
  publisher={Nature Publishing Group UK London}
}

@inproceedings{kang2023benchmarking,
  title={Benchmarking self-supervised learning on diverse pathology datasets},
  author={Kang, Mingu and Song, Heon and Park, Seonwook and Yoo, Donggeun and Pereira, S{\'e}rgio},
  booktitle={Proceedings of the IEEE/CVF Conference on Computer Vision and Pattern Recognition},
  pages={3344--3354},
  year={2023}
}

@article{moco-improved,
  author       = {Xinlei Chen and
                  Haoqi Fan and
                  Ross B. Girshick and
                  Kaiming He},
  title        = {Improved Baselines with Momentum Contrastive Learning},
  journal      = {CoRR},
  volume       = {abs/2003.04297},
  year         = {2020},
  url          = {https://arxiv.org/abs/2003.04297},
  eprinttype    = {arXiv},
  eprint       = {2003.04297},
  }

@inproceedings{chen2020simple,
  title={A simple framework for contrastive learning of visual representations},
  author={Chen, Ting and Kornblith, Simon and Norouzi, Mohammad and Hinton, Geoffrey},
  booktitle={International conference on machine learning},
  pages={1597--1607},
  year={2020},
  organization={PMLR}
}

@inproceedings{grill2020bootstrap,
 author = {Grill, Jean-Bastien and Strub, Florian and Altch\'{e}, Florent and Tallec, Corentin and Richemond, Pierre and Buchatskaya, Elena and Doersch, Carl and Avila Pires, Bernardo and Guo, Zhaohan and Gheshlaghi Azar, Mohammad and Piot, Bilal and kavukcuoglu, koray and Munos, Remi and Valko, Michal},
 booktitle = {Proceedings of Advances in Neural Information Processing Systems},
 pages = {21271--21284},
  title = {Bootstrap Your Own Latent - A New Approach to Self-Supervised Learning},
 volume = {33},
 year = {2020}
}

@inproceedings{he2020momentum,
  title={Momentum contrast for unsupervised visual representation learning},
  author={He, Kaiming and Fan, Haoqi and Wu, Yuxin and Xie, Saining and Girshick, Ross},
  booktitle={Proceedings of the IEEE/CVF Conference on Computer Vision and Pattern Recognition},
  pages={9729--9738},
  year={2020}
}

@article{khosla2020supervised,
  title={Supervised contrastive learning},
  author={Khosla, Pranay and Teterwak, Piotr and Wang, Chen and Sarna, Aaron and Tian, Yonglong and Isola, Phillip and Maschinot, Aaron and Liu, Ce and Krishnan, Dilip},
  journal={Advances in Neural Information Processing Systems},
  volume={33},
  pages={18661--18673},
  year={2020},
  note={\url{http://www.deeplearningbook.org}}
}

@inproceedings{ren2022simple,
  title={A simple data mixing prior for improving self-supervised learning},
  author={Ren, Sucheng and Wang, Huiyu and Gao, Zhengqi and He, Shengfeng and Yuille, Alan and Zhou, Yuyin and Xie, Cihang},
  booktitle={Proceedings of the IEEE/CVF Conference on Computer Vision and Pattern Recognition},
  pages={14595--14604},
  year={2022}
}

@article{yang2020rethinking,
  title={Rethinking the value of labels for improving class-imbalanced learning},
  author={Yang, Yuzhe and Xu, Zhi},
  journal={Advances in neural information processing systems},
  volume={33},
  pages={19290--19301},
  year={2020}
}

@inproceedings{kukleva2023temperature,
  author       = {Anna Kukleva and
                  Moritz B{\"{o}}hle and
                  Bernt Schiele and
                  Hilde Kuehne and
                  Christian Rupprecht},
  title        = {Temperature Schedules for self-supervised contrastive methods on long-tail
                  data},
  booktitle    = {The Eleventh International Conference on Learning Representations,
                  {ICLR} 2023, Kigali, Rwanda, May 1-5, 2023},
  publisher    = {OpenReview.net},
  year         = {2023},
  url          = {https://openreview.net/forum?id=ejHUr4nfHhD},
  bibsource    = {dblp computer science bibliography, https://dblp.org}
}

@inproceedings{liu2021self,
  author       = {Hong Liu and
                  Jeff Z. HaoChen and
                  Adrien Gaidon and
                  Tengyu Ma},
  title        = {Self-supervised Learning is More Robust to Dataset Imbalance},
  booktitle    = {The Tenth International Conference on Learning Representations, {ICLR}
                  2022, Virtual Event, April 25-29, 2022},
  publisher    = {OpenReview.net},
  year         = {2022},
  url          = {https://openreview.net/forum?id=4AZz9osqrar},
  bibsource    = {dblp computer science bibliography, https://dblp.org}
}

@inproceedings{bai2023effectiveness,
  author={Jianhong Bai and Zuozhu Liu and Hualiang Wang and Jin Hao and Yang Feng and Huanpeng Chu and Haoji Hu},
  title={On the Effectiveness of Out-of-Distribution Data in Self-Supervised Long-Tail Learning},
  year={2023},
  cdate={1672531200000},
  url={https://openreview.net/forum?id=v8JIQdiN9Sh},
  booktitle={ICLR}}

@inproceedings{jiang2021self,
  title={Self-damaging contrastive learning},
  author={Jiang, Ziyu and Chen, Tianlong and Mortazavi, Bobak J and Wang, Zhangyang},
  booktitle={International Conference on Machine Learning},
  pages={4927--4939},
  year={2021},
  organization={PMLR}
}

@article{krizhevsky2009learning,
  title={Learning multiple layers of features from tiny images},
  author={Krizhevsky, Alex and Hinton, Geoffrey and others},
  year={2009},
  publisher={Toronto, ON, Canada}
}

@article{russakovsky2015imagenet,
  title={Imagenet large scale visual recognition challenge},
  author={Russakovsky, Olga and Deng, Jia and Su, Hao and Krause, Jonathan and Satheesh, Sanjeev and Ma, Sean and Huang, Zhiheng and Karpathy, Andrej and Khosla, Aditya and Bernstein, Michael and others},
  journal={International journal of computer vision},
  volume={115},
  pages={211--252},
  year={2015},
  publisher={Springer}
}

@inproceedings{azizi2021big,
  author = {Azizi, Shekoofeh and Mustafa, Basil and Ryan, Fiona and ... and Norouzi, Mohammad},
  title = {Big Self-Supervised Models Advance Medical Image Classification},
  booktitle = {IEEE/CVF International Conference on Computer Vision (ICCV)},
  year = {2021}
}

@article{wolf2023self,
  title={Self-supervised pre-training with contrastive and masked autoencoder methods for dealing with small datasets in deep learning for medical imaging},
  author={Wolf, Daniel and Payer, Tristan and Lisson, Catharina Silvia and Lisson, Christoph Gerhard and Beer, Meinrad and G{\"o}tz, Michael and Ropinski, Timo},
  journal={Scientific Reports},
  volume={13},
  number={1},
  pages={20260},
  year={2023},
  publisher={Nature Publishing Group UK London}
}

@inproceedings{tian2021divide,
  title={Divide and contrast: Self-supervised learning from uncurated data},
  author={Tian, Yonglong and Henaff, Olivier J and Van den Oord, A{\"a}ron},
  booktitle={Proceedings of the IEEE/CVF International Conference on Computer Vision},
  pages={10063--10074},
  year={2021}
}

@article{assran2022hidden,
  title={The hidden uniform cluster prior in self-supervised learning},
  author={Assran, Mahmoud and Balestriero, Randall and Duval, Quentin and Bordes, Florian and Misra, Ishan and Bojanowski, Piotr and Vincent, Pascal and Rabbat, Michael and Ballas, Nicolas},
  journal={arXiv preprint arXiv:2210.07277},
  year={2022}
}

@inproceedings{lin2023frequency,
  title={Frequency-aware self-supervised long-tailed learning},
  author={Lin, Ci-Siang and Chen, Min-Hung and Wang, Yu-Chiang Frank},
  booktitle={Proceedings of the IEEE/CVF International Conference on Computer Vision},
  pages={963--972},
  year={2023}
}

@article{sharma2025maximally,
  title={Maximally Useful and Minimally Redundant: The Key to Self Supervised Learning for Imbalanced Data},
  author={Sharma, Yash Kumar and Nair, Vineet and Naik, Wilson},
  journal={arXiv preprint arXiv:2509.08469},
  year={2025}
}

@article{espis2025comparative,
  title={Comparative analysis of supervised and self-supervised learning with small and imbalanced medical imaging datasets},
  author={Espis, Andrea and Marzi, Chiara and Diciotti, Stefano},
  journal={Scientific Reports},
  volume={15},
  number={1},
  pages={32345},
  year={2025},
  publisher={Nature Publishing Group UK London}
}
\end{document}